\pdfoutput=1

\documentclass[11pt]{article}

\usepackage[final]{acl}

\usepackage{times}
\usepackage{latexsym}
\usepackage{bm}
\usepackage{amsmath}
\usepackage{amssymb}
\usepackage{mathtools}
\usepackage{amsthm}
\usepackage{multirow}
\usepackage{lscape}
\usepackage{algorithm}
\usepackage{algpseudocode}
\usepackage[capitalize,noabbrev]{cleveref}
\usepackage{arydshln}
\theoremstyle{plain}
\newtheorem{theorem}{Theorem}[section]

\newtheorem{corollary}[theorem]{Corollary}
\theoremstyle{definition}

\theoremstyle{remark}

\usepackage[textsize=tiny]{todonotes}
\usepackage[inkscapelatex=false]{svg}
\usepackage{subfigure}
\usepackage[T1]{fontenc}

\usepackage[utf8]{inputenc}
\usepackage{enumitem}
\usepackage{microtype}
\usepackage{pifont}

\usepackage{graphicx}

\usepackage{tikz}
\newcommand*\circled[1]{\tikz[baseline=(char.base)]{
            \node[shape=circle,draw,inner sep=1pt] (char) {#1};}}
\title{FLAT-LLM: Fine-grained Low-rank Activation Space Transformation for \\ Large Language Model Compression}

\author{First Author \\
  Affiliation / Address line 1 \\
  Affiliation / Address line 2 \\
  Affiliation / Address line 3 \\
  \texttt{email@domain} \\\And
  Second Author \\
  Affiliation / Address line 1 \\
  Affiliation / Address line 2 \\
  Affiliation / Address line 3 \\
  \texttt{email@domain} \\}

\author{
 \textbf{Jiayi Tian\textsuperscript{1}},
 \textbf{Ryan Solgi\textsuperscript{1}},
 \textbf{Jinming Lu\textsuperscript{1}},
 \textbf{Yifan Yang\textsuperscript{1}},
 \textbf{Hai Li\textsuperscript{2}},
 \textbf{Zheng Zhang\textsuperscript{1}} \\
\\
 \textsuperscript{1} University of California, Santa Barbara,
 \textsuperscript{2} Intel Corporation
\\
 \small{
   {\{jiayi\_tian, solgi, jinminglu, yifanyang\}@ucsb.edu},
   {hai.li@intel.com},
   {zhengzhang@ece.ucsb.edu}
 }
}

\begin{document}
\maketitle
\begin{abstract}
Large Language Models (LLMs) have enabled remarkable progress in natural language processing, yet their high computational and memory demands pose challenges for deployment in resource-constrained environments. 
Although recent low-rank decomposition methods offer a promising path for structural compression, they often suffer from accuracy degradation, expensive calibration procedures, and result in inefficient model architectures that hinder real-world inference speedups.
In this paper, we propose FLAT-LLM, a fast and accurate, training-free structural compression method based on fine-grained low-rank transformations in the activation space. 
Specifically, we reduce the hidden dimension by transforming the weights using truncated eigenvectors computed via head-wise Principal Component Analysis, and employ a greedy budget redistribution strategy to adaptively allocate ranks across decoders.
FLAT-LLM achieves efficient and effective weight compression without recovery fine-tuning, which could complete the calibration within a few minutes.  
Evaluated across 5 models and 11 datasets, FLAT-LLM outperforms structural pruning baselines in generalization and downstream performance, while delivering inference speedups over low-rank-based methods. 
\footnote{Code is available in \url{https://github.com/TTTTTTris/FLAT-LLM}.}

\end{abstract}

\section{Introduction}
Large Language Models (LLMs) have achieved state-of-the-art performance in a wide range of natural language processing and understanding tasks \cite{bai2023qwen, liu2024deepseek, touvron2023llama}.
However, their substantial parameter counts and computational demands pose significant challenges for deployment in edge devices and resource-constrained environments.
Model compression is a promising direction for reducing both the model size and computational complexity, with popular approaches including quantization \cite{huang2024billm, tian2023bebert}, knowledge distillation \cite{sun2020mobilebert, jiao2020tinybert}, pruning \cite{sunsimple, ma2023llm, yang2025wanda}, and low-rank decomposition \cite{ashkboosslicegpt, yang2024loretta}. 
Among these, low-rank decomposition stands out for its hardware efficiency due to its structural nature.

\begin{figure}[t]
    \centering
    \includegraphics[width=\linewidth]{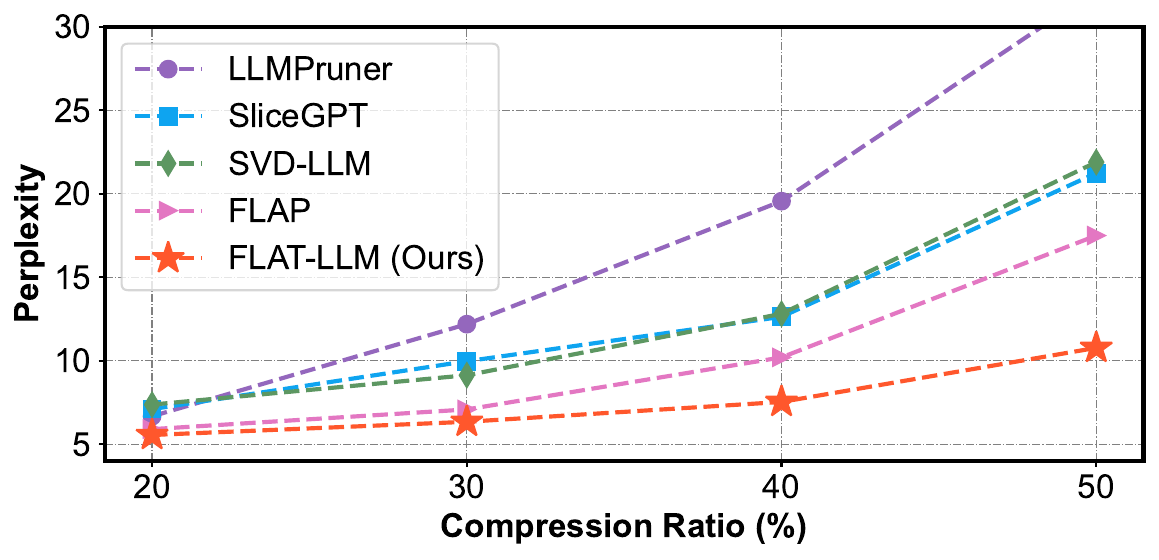}
    \caption{Comparison of WikiText-2 perplexity against various baselines on Llama-2 13B model.}
    \label{fig:ppl}
\end{figure}

\begin{figure*}[htbp]
    \centering
    \includegraphics[width=\linewidth]{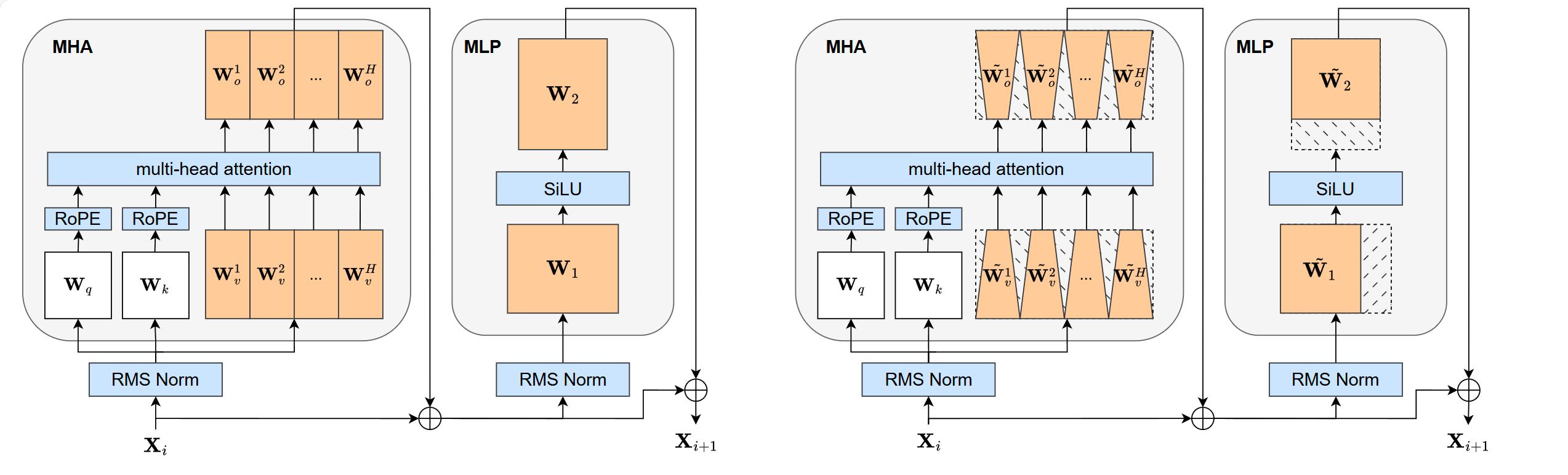}
    \caption{Decoder structure before (left) and after (right) weight truncation. Orange blocks indicate truncated weights; hatched areas show removed weights; blue boxes denote non-linear functions.}
    \label{fig:1}
\end{figure*}

Singular Value Decomposition (SVD) is a common low-rank decomposition approach for compressing the weight matrices of LLMs. However, weights from standard pre-training are often nearly full-rank and difficult to compress \cite{yu2023compressing}. 
To address this, recent work uses the low-rank nature of activation spaces by projecting weights into these subspaces to improve the compression ratio. 
For example, ASVD \cite{yuan2023asvd} uses activation norms to transform the weights before decomposition, and SVD-LLM \cite{wang2024svd} enhances the accuracy by linking truncated singular values to reconstruction loss.

Despite recent improvements, SVD-based methods often incur performance drops and require recovery fine-tuning, even under low compression ratios. 
This stems from a key limitation: preserving sufficient information via high-rank SVD increases model parameters, as both left and right singular vectors must be stored. 
The problem is especially severe for square matrices (which are common in LLMs like Llama), where reducing parameters requires truncating at least 50\% of singular values, often leading to significant information loss.
Moreover, replacing a single large matrix multiplication with two smaller ones in SVD-format linear layers can degrade GPU efficiency, limiting practical inference speedups.

To address the inherent limitations of SVD, SliceGPT \cite{ashkboosslicegpt} projects post-RMSNorm hidden states into low-rank subspaces using Principal Component Analysis (PCA), enabling the resulting eigenvector pairs to be fully absorbed into adjacent layers.
However, to enable this transformation between decoder layers, SliceGPT requires inserting adapter modules along the residual paths, which bring high memory overhead and limit the inference speedup. 
For example, on Llama-2 7B, SliceGPT incur 10\% additional memory overhead at a 20\% compression ratio, and yields only $1.1\times$ speedup in inference at a 50\% compression ratio.

To further enhance the performance of low-rank compressed models, prior works have introduced rank-adaptive methods that allocate heterogeneous ranks across different model components. For example, Adaptive SVD~\cite{gao2024adaptive} employs a trainable hypernetwork to determine layer-wise ranks, achieving higher compression efficiency than uniform baselines but at the cost of task-specific retraining. However, this approach requires weeks on LLaMA-7B, making it impractical and highly inefficient for scaling to larger models such as LLaMA-70B. 
 \textcolor{black}{MoDeGPT~\cite{lin2024modegpt} adopts an importance-score-based sparsity allocation scheme with entropic regularization for smoothing, but it demands extensive hyperparameter tuning to obtain competitive performance.}



To overcome the challenges mentioned above, we propose FLAT-LLM, a fast, accurate, and training-free structural compression method for LLMs.
Figure~\ref{fig:1} illustrates the decoder architecture before and after applying FLAT-LLM compression.
FLAT-LLM projects the post-value hidden states into low-rank subspaces using fine-grained, head-wise PCA. The resulting low-rank eigenvector pairs are then absorbed into the value and output weight matrices to complete the compression.
\textcolor{black}{Unlike previous approaches, this joint absorption-based compression introduces no additional memory overhead, and the overall compression ratio directly corresponds to the retained rank ratio.}
To further improve performance, we introduce an importance-preserving rank selection algorithm.
This algorithm is entirely tuning-free and completes within seconds, achieving over 100$\times$ higher time efficiency compared to Adaptive SVD~\cite{gao2024adaptive}, \textcolor{black}{and consistently outperforming MoDeGPT’s rank selection strategy.}
Our main contributions are as follows:
\begin{itemize}[leftmargin=*, labelsep=0.5em]
    \item We propose a training-free, fine-grained compression technique that operates within multi-head attention layers, avoiding the inefficiencies of prior decomposition-based methods.
    \item We introduce a novel training-free rank selection algorithm that allocates ranks using a greedy redistribution strategy and can be integrated with existing low-rank LLM compression pipelines.
    \item 
    We demonstrate the effectiveness of FLAT-LLM through extensive evaluations on language modeling and downstream tasks. 
    As shown in Figure~\ref{fig:ppl}, FLAT-LLM significantly improves perplexity on the WikiText-2 test split across a range of compression ratios, indicating enhanced text generation capabilities under varying levels of compression.
\end{itemize}

\section{Related Work}
In addition to structural pruning via weight decomposition (as discussed in the introduction), another line of research achieves structural compression by directly removing model components based on importance scores. 
These approaches can be broadly categorized into two types: fine-grained pruning and coarse-grained pruning. 
The former offers fine-grained control by removing individual rows or columns within weight matrices. For example, LLM-Pruner \cite{ma2023llm} leverages gradient-based saliency scores to prune less important components, while FLAP \cite{an2024fluctuation} adaptively removes unstable neurons and channels based on activation fluctuations. 

In contrast, the coarse-grained pruning eliminates larger components such as attention heads, layers, or entire decoders. 
Though more efficient for inference, this often results in greater performance degradation. For instance, ShortGPT \cite{men2024shortgpt} prunes decoders based on cosine similarity-based importance ranking; LaCo \cite{yang2024laco} compresses models by merging adjacent layers; and BlockPruner \cite{zhong2024blockpruner} removes redundant MHA or MLP blocks through iterative search. 
In this work, we demonstrate that our method outperforms both fine- and coarse-grained structural pruning baselines in terms of performance under compression.

\section{Background}
\paragraph{Multi-Head Attention.}
A typical LLM consists of multiple decoder layers, each comprising two main components: Multi-Head Attention (MHA) and a Multi-Layer Perceptron (MLP). We define the input hidden states \( \mathbf{X} \in \mathbb{R}^{N \times d_{hid}} \), where \(N\) denote the sequence length, and \(d_{hid}, d_h\) represent the dimension of hidden states and each attention head, respectively. Here, the MHA module computes a weighted aggregation of values using learned projections of queries, keys, and values. For a total of \( H \) attention heads, each head \( h \) performs:
\begin{align}
\begin{split}
\mathbf{A}^h &= \frac{ \mathbf{X}{\mathbf{W}_q^h}{^\top} (\mathbf{X}{\mathbf{W}_k^h}{^\top}){^\top} }{ \sqrt{d_h} }, \\
\mathbf{Y}_v^h &= \text{Softmax}(\mathbf{A}^h)\mathbf{X} {\mathbf{W}_v^h}{^\top}, 
\label{eqa:1}
\end{split}
\end{align}
where \( \mathbf{W}_q^h, \mathbf{W}_k^h, \mathbf{W}_v^h \in \mathbb{R}^{d_h \times d_{hid}} \) are the projection matrices in query, key and value layer in head $h$. The attention matrix \(\mathbf{A}^h \in \mathbb{R}^{N \times N}\) captures token-wise interactions, and the Softmax function is applied to compute attention scores, which are then used to weight the value representations. The result \( \mathbf{Y}_v^h \in \mathbb{R}^{N \times d_h} \) represents the per-head value output. 
This is is further transformed by a learned output projection \( \mathbf{W}_o^h \in \mathbb{R}^{d_{hid} \times d_{h}} \), yielding the partial attention output \( \mathbf{Y}_o^h \in \mathbb{R}^{N \times d_{hid}}\). 
The final output of the multi-head attention layer is then obtained by aggregating the partial outputs from all heads:
\begin{align}
\begin{split}
\mathbf{Y}_o^h &=  \text{Softmax}(\mathbf{A}^h)\mathbf{X} {\mathbf{W}_v^h}{^\top}{\mathbf{W}_o^h}{^\top}, \\
\mathbf{Y}_o &= \text{sum}(\mathbf{Y}_o^1, \dots, \mathbf{Y}_o^H).
\label{eqa:2}
\end{split}
\end{align}

\paragraph{Principle Component Analysis (PCA).}
\textcolor{black}{Principal Component Analysis (PCA) is a classical dimensionality reduction technique that identifies the the principal components along which the data exhibit the greatest variance. 
Given a data matrix \( \mathbf{Z} \in \mathbb{R}^{N \times d} \), PCA computes the eigen decomposition of the covariance matrix $\mathbf{C}$:
\begin{align}
\mathbf{C} = {\mathbf{Z}}{^\top} \mathbf{Z} = \mathbf{Q} \mathbf{\Lambda}\mathbf{Q}{^\top},
\label{eqa:3}
\end{align}
where \( \mathbf{Q} \in \mathbb{R}^{d \times d} \) is the orthogonal matrix of eigenvectors, and \( \mathbf{\Lambda} \in \mathbb{R}^{d \times d} \) is the diagonal matrix of corresponding eigenvalues. 
To capture the most significant variance, we retain the top-\( r \) principal components and define \( \tilde{\mathbf{Q}} \in \mathbb{R}^{d \times r} \) as the truncated eigenvector matrix. 
The corresponding rank-$r$ reconstruction of the original data is defined as projecting $\mathbf{Z}$ onto the rank-$r$ space and then mapping it back to the original space:
\begin{align}
\tilde{\mathbf{Z}} = \mathbf{Z} \tilde{\mathbf{Q}}\tilde{\mathbf{Q}}^T.
\end{align}
PCA thus reveals low-rank structure in weight matrices or hidden representations, enabling efficient approximations with minimal loss of information.}

\section{FLAT-LLM}

In this section, we present FLAT-LLM in detail, beginning with a head-wise PCA-based weight truncation method that compresses weight matrices by truncating and absorbing the eigenvectors. 
We then introduce an importance-preserving rank selection strategy to allocate adaptive ranks across decoder layers.
Finally, we conduct a theoretical analysis of the truncation loss in our head-wise PCA method.
\begin{figure}
    \centering
    \includegraphics[width=\linewidth]{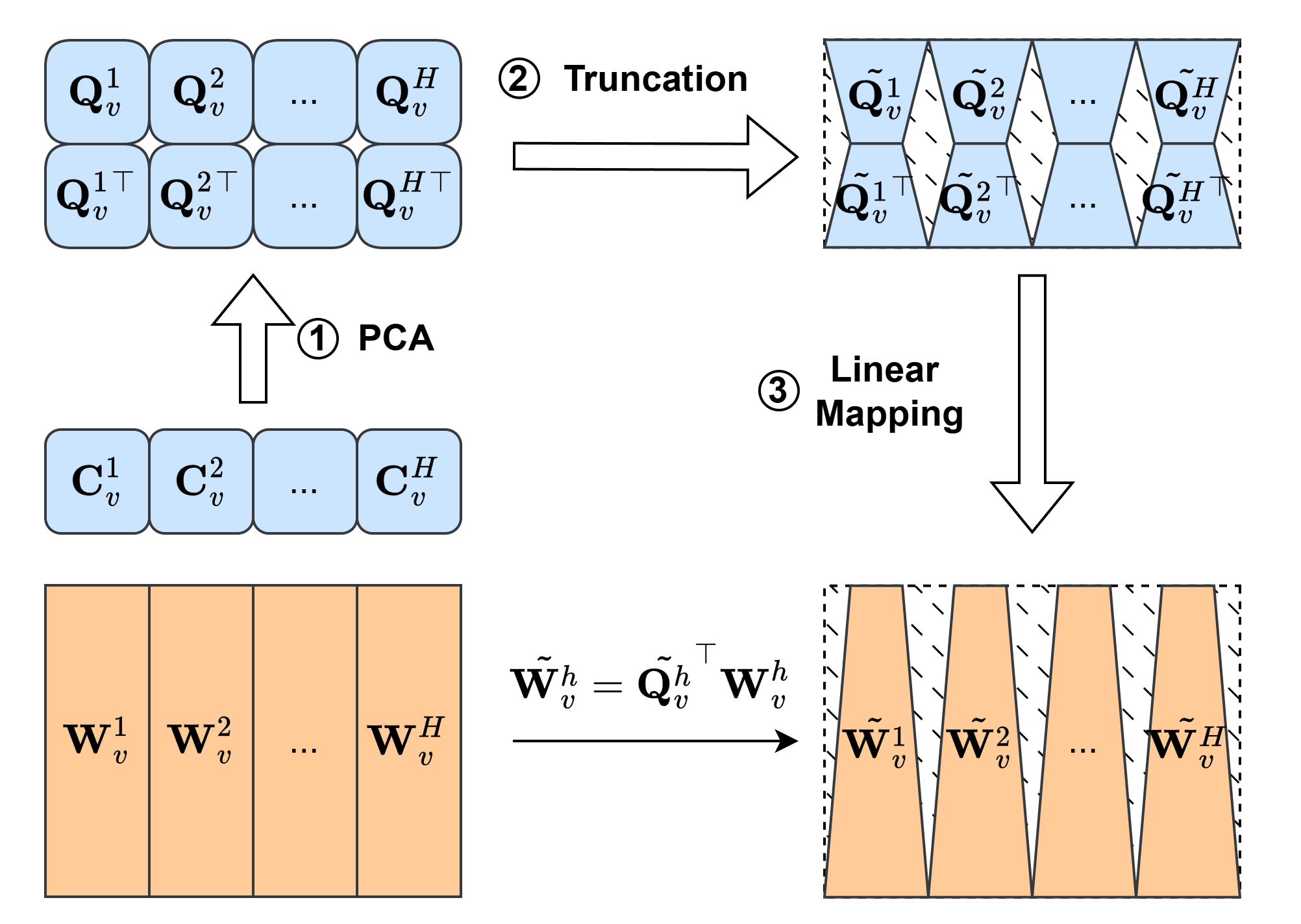}
    \caption{Fine-grained head-wise PCA in value layer.}
    \label{fig:ATTN}
\end{figure}

\subsection{Head-wise PCA-based Weight Truncation}
\label{sec:head}
\textcolor{black}{Inspired by Equation~\eqref{eqa:2}, we observe that the value and output projections within each attention head are computed consecutively. Leveraging this observation, we propose to exploit the low-rank activation space of the value output to jointly compress the value and output weight matrices.}

As shown in Figure \ref{fig:ATTN}, the detail compression process with $M$ calibration samples are given in three steps:
\circled{1} Compute the covariance matrix $\mathbf{C}_v^h = \sum_{m=1}^M {\mathbf{Y}^h_{v,m}}{^\top} {\mathbf{Y}^h_{v,m}}$ and perform PCA to obtain the orthogonal eigenvector $\mathbf{Q}_v^h \in \mathbb{R}^{d_h\times d_{h}}$ utilizing Equation \eqref{eqa:3}.
\circled{2} Truncate the eigenvectors to rank $r$, yielding reduced basis $\tilde{\mathbf{Q}}_v^h \in \mathbb{R}^{d_h\times r}$, and the reconstructed per-head value output becomes $\tilde{\mathbf{Y}}^h_{v} = \mathbf{Y}^h_{v} \tilde{\mathbf{Q}}_v^h\tilde{\mathbf{Q}}_v^h{^\top}.$ \circled{3} To compress the weights by absorbing the truncated eigenvectors, we reformulate the MHA computation in Equation \eqref{eqa:2} as the following:
\begin{align}
&\mathbf{Y}_o^h =  \text{Softmax}(\mathbf{A}^h)\mathbf{X} {\mathbf{W}_v^h}{^\top}{\mathbf{Q}_v^{h}}{\mathbf{Q}_v^{h}}{^\top}{\mathbf{W}_o^h}{^\top}
\label{eqa:7}
\end{align}
This enables the jointly compression on the value and output weights using the PCA basis derived from the value layer:
\begin{align}
\begin{split}
&\tilde{\mathbf{Y}}_{o}^{h} = \text{Softmax}(\mathbf{A}^{h})\mathbf{X}\mathbf{W}_v^{h}{^\top}\tilde{\mathbf{Q}}_v^{h}\tilde{\mathbf{Q}}_v^{h}{^\top}\mathbf{W}_o^{h}{^\top},  \\
&\tilde{\mathbf{Y}}_{o}^{h} = \text{Softmax}(\mathbf{A}^{h})\mathbf{X}\tilde{\mathbf{W}}_v^{h}{^\top}\tilde{\mathbf{W}}_o^{h}{^\top}, 
\label{eqa:8}
\end{split}
\end{align}
where the first and second equations represent the truncation and absorption.
Here, we aim for $\tilde{\mathbf{Q}}_v^h {\tilde{\mathbf{Q}}_v^h}{^\top} \approx \mathbf{I}_v^h$ to retain most of the representational power of the original $\mathbf{Y}_v^h$. 
After absorbing the truncated basis into the weights, the value and output projections of each head are reduced to $\tilde{\mathbf{W}}_v^h \in \mathbb{R}^{r\times d_{hid}}$, $\tilde{\mathbf{W}}_o^h \in \mathbb{R}^{d_{hid}\times r}$, respectively. 

In this way, we can jointly compress both the value and output weights to $\frac{r}{d_h}$ of their original size leveraging the low-rank structure of the output hidden states from the value layer. 
Notably, this joint compression technique remains compatible with modern architectures using Grouped-Query Attention (GQA) \cite{ainslie2023gqa}, resulting in different numbers of value and output heads. 
The detailed formulation for the GQA case is provided in Appendix~\ref{appendix:GQA}.

Additionally, although the query and key projections cannot be jointly compressed in the same manner as the value and output weight matrices, head-wise PCA can still be applied to them independently. 
Specifically, we follow the same three-step head-wise PCA procedure illustrated in Figure~\ref{fig:ATTN}.
The reconstructed query and key outputs for head $h$ can be written as
\begin{equation}
\tilde{\mathbf{Y}}^h_{q} = \mathbf{Y}^h_{q} \tilde{\mathbf{Q}}_q^h\tilde{\mathbf{Q}}_q^h{^\top}, 
\tilde{\mathbf{Y}}^h_{k} = \mathbf{Y}^h_{k} \tilde{\mathbf{Q}}_k^h\tilde{\mathbf{Q}}_k^h{^\top}.
\end{equation}
Accordingly, the compressed query and key projection matrices are given by
\begin{equation}
\tilde{\mathbf{W}}^h_q = \tilde{\mathbf{Q}}^h_q \mathbf{W}^{h\top}_q,
\qquad
\tilde{\mathbf{W}}^h_k = \tilde{\mathbf{Q}}^h_k \mathbf{W}^{h\top}_k .
\end{equation}
Since the reduced basis $\tilde{\mathbf{Q}}_v^h \in \mathbb{R}^{d_h\times r}$ is $H\times$ is significantly smaller than the corresponding compressed projection matrices $\tilde{\mathbf{W}}^h_{q,k} \in \mathbb{R}^{r \times d_{\mathrm{hid}}}$, performing head-wise PCA on the query and key projections leads to substantially lower approximation error than applying a full-matrix low-rank decomposition at the same sparsity level.

\begin{algorithm}[t]
\caption{Importance-Preserving Rank Selection}
\label{alg:1}
\begin{algorithmic}[1]
\Require Sparsity $s$, number of decoders $L$, importance scores $\mathbf{t}$
\State Initialize total budget $B \leftarrow L(1-s)$, active set $\mathcal{A} \leftarrow \{1, \dots, L\}$ 
\While{$\mathcal{A} \neq \emptyset$}
    \State $\tilde{w}_l = \frac{t_l}{\sum_{j \in \mathcal{A}} t_j} \cdot B$ for all $l \in \mathcal{A}$ 
    \State Let $\mathcal{S} \leftarrow \{l \in \mathcal{A} \mid \tilde{w}_l > 1\}$ 
    \If{$\mathcal{S} = \emptyset$} \Comment{Assign all remain entries}
        \State $w_l \leftarrow \tilde{w}_l$ for all $l \in \mathcal{A}$ 
        \State \textbf{break} 
    \EndIf
   \ForAll{$l \in \mathcal{S}$}
    \Comment{Assign fixed entries}
        \State $w_l \leftarrow 1$ 
        \State $B \leftarrow B - w_l$ 
    \EndFor
    \State $\mathcal{A} \leftarrow \mathcal{A} \setminus \mathcal{S}$  \Comment{Remove fixed entries}
\EndWhile
\State \Return $\mathbf{w}$ \Comment{Final allocation}
\end{algorithmic}
\end{algorithm}

\subsection{Importance-Preserving Rank Selection}

\textcolor{black}{
In our experiments, we observed that using a uniform rank across all layers degrades performance, particularly under high compression ratios. 
We first analyze the cosine similarity between each decoder layer’s input and output hidden states, and reveal that the intrinsic dimensionality varies across layers. 
Motivated by this observation, we propose a decoder-wise rank selection algorithm that employs a greedy redistribution strategy to adaptively allocate ranks based on their relative importance.
}

\textcolor{black}{
To analyze variations in intrinsic dimension across decoder layers, we compute the cosine similarity between the input and output hidden states of each decoder.
}
Given a model with $L$ decoder layers, let \( \mathbf{X}_{l} \) and \( \mathbf{X}_{l+1} \) denote the input and output hidden state matrices of the $l$-th decoder layer, respectively. For each sample \( p \in \{1, \dots, N\} \), the cosine similarity \( c_{l,p} \) between the corresponding rows of \( \mathbf{X}_{l} \) and \( \mathbf{X}_{l+1} \) is defined as:
\begin{align}
c_{l,p} = \frac{\mathbf{X}_{l,p}{^\top} \mathbf{X}_{l+1,p}}{\|\mathbf{X}_{l,p}\|_2 \, \|\mathbf{X}_{l+1,p}\|_2}, 
\end{align}
where \( \mathbf{X}_{l,p} \) and \( \mathbf{X}_{l+1,p} \) represent the $p$-th row of \( \mathbf{X}_{l} \) and \( \mathbf{X}_{l+1} \), respectively. The average cosine similarity for the $l$-th decoder layer is then given by \( c_l = \mathbb{E}_{X,p}[c_{l,p}] \), reflecting the overall alignment between input and output hidden states.

\textcolor{black}{
To quantify the degree of dissimilarity and thereby infer the intrinsic dimension, we compute $t_l = \frac{\arccos{c_l}}{\pi},$ 
which captures the normalized angular deviation between the representations. 
We interpret \( t_l \) as an indicator of the relative importance and compressibility of each decoder layer. From empirical evaluations on several LLMs, we observe that \( t_l \) typically ranges from 0.06 to 0.30. This leads to two key insights: (1) intrinsic dimensionality varies across layers, motivating the use of heterogeneous rank assignments; and (2) the small angular deviations suggest that decoder layers often reside in low-dimensional subspaces, indicating that LLMs are amenable to compression.
}


Given the model sparsity as $s$, to determine the remaining rank ratios $w_l$ for $l$-th decoder regarding the importance score $t_l$, a naive way is to proportional scale the total budget of remaining rank ratios $B=L(1-s)$, which gives $\hat{w}_l = \frac{t_l}{\sum_{l=1}^L t_l} B$. However, this naive proportional allocation may violate the $w_l \in[0,1]$ constraint when some components receive disproportionately high scores. In order to design $w_l$ regarding the importance score $t_l$, while fit each $w_l$ within the constraint and fix the total budget for remaining size as $\sum_{l}w_l=B$, the objective of our rank selection algorithm can be defined as:
\begin{align}
\text{min}_{\mathbf{w}\in[0,1]^L}||\mathbf{w}-\mathbf{\hat{w}}||,\, \text{s.t.} \sum_{l}w_l=B,
\end{align}
where $\mathbf{w}=[w_1, \cdots, w_L]$ and $\mathbf{\hat{w}}=[\hat{w}_1, \cdots,\hat{w}_L]$ are vectors constructed by the ratios $w_l$ and $\hat{w}_l$ in each decoders.

To address this, we implement a greedy redistribution strategy, as shown in Algorithm \ref{alg:1}. First, we define a variable $\tilde{w}_l$ to represent the proportional scaled ratios $\hat{w}_l$ when the budget $B$ is changing during the rank selection process. Given an active set $\mathcal{A}$ that contains the indices of unassigned remaining rank ratios, we iteratively compute $\tilde{w}_l$ in the active set with latest budget $B$, clip $\tilde{w}_l$ that exceed the upper bound as 1, update the total budget $B$, and remove the clipped entries from the active set. 
This process continues until all elements has been assigned a remaining rank ratios $w_l$. 
In this way, the resulting solution $w_l$ remains proportional to $t_l$ where possible, while ensuring the boundedness and budget feasibility.

\begin{figure}
    \centering
    \includegraphics[width=\linewidth]{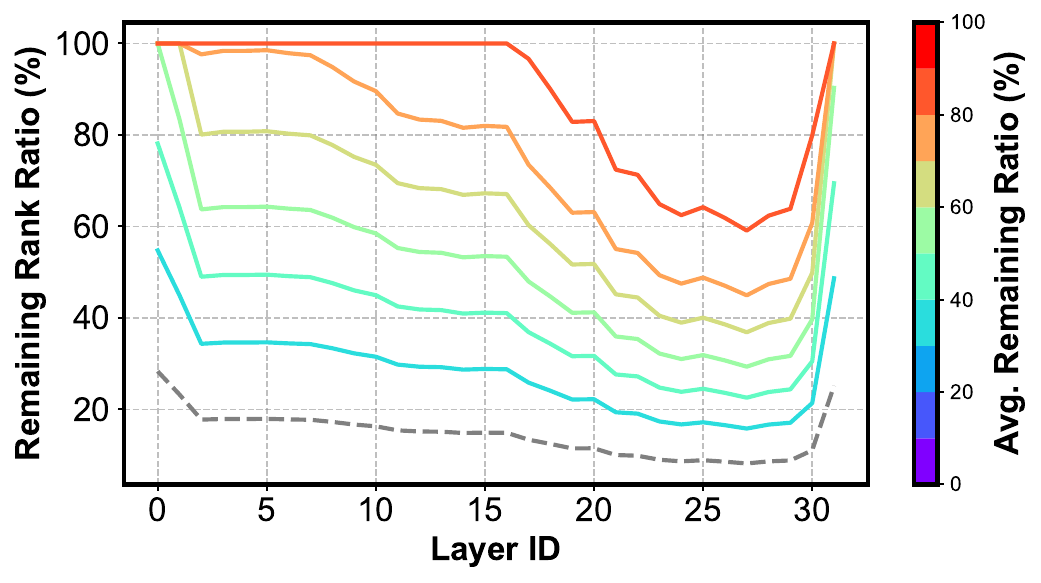}
    \caption{Remaining rank ratio versus layer id computed with Algorithm \ref{alg:1}. The average remaining ratio is set between 30\% (lowest solid) to 90\% (highest solid). }
    \label{fig:IPRS}
\end{figure}


Figure~\ref{fig:IPRS} illustrates the remaining rank ratios across the decoder blocks of Llama-2 7B under different average remaining ratios (from 40\% to 90\%) determined by the IPRS algorithm. The gray dashed curve represents the normalized importance scores $\mathbf{t}$ that guide the optimization process, while the colorful solid lines show the remaining rank ratio $\mathbf{w}$ under multiple sparsity ratio $s \in [0.1,0.6]$. 
As shown, the resulting rank allocations are non-uniform and adaptively preserve more capacity in layers with higher importance. Additional visualizations of the rank distributions produced by IPRS on other models are provided in Appendix~\ref{appendix:IPRS}.

\subsection{Truncation Error Analysis}
In the following, we provide the theoretical proof of the direct correspondence between eigenvalues and truncation loss in both single-head and multi-head cases, which guarantees the effectiveness of our method. 
The detailed derivations are presented in Appendix~\ref{appendix:proof}. \textcolor{black}{We also include an empirical comparison of reconstruction errors in Appendix~\ref{appendix:emprical_error} with SVD-LLM to further demonstrate the advantage of our head-wise PCA for the attention block.}

As described in Section \ref{sec:head}, to compress the value layer, we perform PCA on its feature covariance and project onto a low-rank subspace:
\begin{align}
&\tilde{\mathbf{Y}}_v^h = \mathbf{Y}_v^h\tilde{\mathbf{Q}}_v^h {\tilde{\mathbf{Q}}_v^h}{^\top}, \\
\text{where } &{\mathbf{Y}_v^h}{^\top}\mathbf{Y}_v^h = \mathbf{Q}_v^h \boldsymbol{\Lambda}_v^h {\mathbf{Q}_v^h}{^\top}, \tilde{\mathbf{Q}}_v^h = \mathbf{Q}_v^h[:,:r]. \nonumber
\end{align}
Here, \( \mathbf{Q}^h \in \mathbb{R}^{d_h \times d_h} \) contains orthonormal eigenvectors and \( \boldsymbol{\Lambda}^h \) is diagonal with eigenvalues \( \lambda_1^h \geq \cdots \geq \lambda_{d_h}^h \geq 0 \), and \( \tilde{\mathbf{Q}}^h \in \mathbb{R}^{d_h \times r} \) contains the top-$r$ principal components.

\begin{theorem}[Reconstruction Error of Single-head Output PCA Projection]
Let \( \mathbf{Y}_v^h =  \mathbf{X} {\mathbf{W}_v^h}{^\top}\), and \( \tilde{\mathbf{Y}}_v^h = \mathbf{Y}_v^h\tilde{\mathbf{Q}}_v^h {\tilde{\mathbf{Q}}_v^h}{^\top} \) be the rank-\( r \) approximation obtained by projecting \( \mathbf{Y}_v^h \) onto its top-\( r \) principal components. Then the squared Frobenius norm of the reconstruction error satisfies:
\vspace{-0.2cm}
\[\| \mathbf{Y}_v^h - \tilde{\mathbf{Y}}_v^h \|_F^2 = \sum_{i = r+1}^{d_h} \lambda_{i}^h,\]
\vspace{-0.2cm}
where \( \{ \lambda_i^h \} \) are the eigenvalues of \( {\mathbf{Y}^h_v}{^\top} \mathbf{Y}_v^h \).
\end{theorem}

\begin{corollary}[Reconstruction Error of Multi-head Output PCA Projection]
Let $\mathbf{Y}_v = \text{concat}(\mathbf{Y}_v^1,...\mathbf{Y}_v^H)$ be the concatenated output hidden states. 
The squared Frobenius norm of the reconstruction error satisfies:
\[
\| \mathbf{Y}_v - \tilde{\mathbf{Y}}_v \|_F^2 = \sum_{h=1}^H\sum_{i = r+1}^{d_h} \lambda_i^h,
\]
\end{corollary}

Therefore, the reconstruction error of the multi-head value output can be expressed as the sum of the truncated eigenvalues $\lambda_i^h$ from the output of each head value projection $\mathbf{Y}_v^h$. 
When the preserved dimension is $r$, truncating the smallest eigenvalues $\lambda_{r+1}^h, .., \lambda_d^h$ in the PCA decomposition of each value head yields the minimal reconstruction error for the multi-head value output.

\begin{table*}[htbp]
\centering
\caption{Comparison of downstream performance against prior structural compression methods on LLaMA-2 (7B, 13B, 70B), LLaMA-3 8B, and Mistral-7B models models at 20\% compression ratio. $\dag$ indicates FLAT-LLM with head-wise query and key compression.}
\resizebox{\textwidth}{!}{
\begin{tabular}{c|c|c|c|ccccccc}
\hline
\textbf{Model}                         & \textbf{Method}                           & {\textbf{PPL} $\downarrow$}      & {\textbf{Avg.} $\uparrow$}  & {\textbf{MMLU 5-shot}} & {\textbf{PIQA}}      & {\textbf{WinoG.}}    & {\textbf{HellaS.}}   & {\textbf{ARC-e}}     & {\textbf{ARC-c}}     & {\textbf{OBQA}}      \\ \hline
                                       & Original                                     & 5.11                                  & 62.14                                  & 45.70                                    & 79.05                                  & 69.38                                  & 75.92                                  & 74.49                                  & 46.25                                  & 44.20                                  \\\cline{2-11}
                                       & LLM-Pruner \cite{ma2023llm}                              & 10.55                                 & 53.89                                  & 26.20                                    & \textbf{75.95}                         & 63.38                                  & 67.83                        & 64.31                                  & 39.93                        & 39.60                                  \\
                                       & FLAP \cite{an2024fluctuation}                                     & 6.76                                  & 53.07                                  & 31.90                                    & 74.54                                  & 62.98                                  & 64.74                                  & 61.28                                  & 36.43                                  & 39.60                                  \\
                                       & SliceGPT \cite{ashkboosslicegpt}                                 & 8.24                                  & 46.26                                  & 26.75                                    & 64.80                                  & 62.98                                  & 49.18                                  & 55.68                                  & 31.40                                  & 33.00                                  \\
                                       & SVD-LLM \cite{wang2024svd}                                  & 7.84                                  & 52.32                                  & 29.34                                    & 71.49                                  & 65.27                                  & 58.57                                  & \textbf{68.31}                         & 35.84                                  & 37.40                                  \\
  & \cellcolor[HTML]{DAE8FC}\textbf{FLAT-LLM (ours)} & \cellcolor[HTML]{DAE8FC}6.70 & \cellcolor[HTML]{DAE8FC}55.16 & \cellcolor[HTML]{DAE8FC}39.67   & \cellcolor[HTML]{DAE8FC}72.20          & \cellcolor[HTML]{DAE8FC}65.82 & \cellcolor[HTML]{DAE8FC}64.72          & \cellcolor[HTML]{DAE8FC}64.44 & \cellcolor[HTML]{DAE8FC}38.65          & \cellcolor[HTML]{DAE8FC}40.60 \\ 
\multirow{-6}{*}{\rotatebox{90}{\textbf{LLaMA-2 7B}}}  & \cellcolor[HTML]{DAE8FC}\textbf{FLAT-LLM$^{\dag}$ (ours)} & \cellcolor[HTML]{DAE8FC}\textbf{6.34} & \cellcolor[HTML]{DAE8FC}\textbf{57.85} & \cellcolor[HTML]{DAE8FC}\textbf{43.99}   & \cellcolor[HTML]{DAE8FC}73.18          & \cellcolor[HTML]{DAE8FC}\textbf{68.35} & \cellcolor[HTML]{DAE8FC}\textbf{68.99}          & \cellcolor[HTML]{DAE8FC}66.79 & \cellcolor[HTML]{DAE8FC}\textbf{40.87}          & \cellcolor[HTML]{DAE8FC}\textbf{42.80} \\ \hline \hline
                                       & Original                                     & 4.57                                  & 65.70                                  & 55.40                                    & 80.41                                  & 72.53                                  & 79.41                                  & 77.39                                  & 49.15                                  & 45.60                                  \\\cline{2-11}
                                       & LLM-Pruner \cite{ma2023llm}                                 & 9.67                                  & 55.45                                  & 22.80                                    & \textbf{77.97}                         & 60.77                                  & 71.26                                  & 67.09                                  & 44.28                                  & \textbf{44.00}                         \\
                                       & FLAP  \cite{an2024fluctuation}                                     & 5.90                                  & 57.00                                  & 41.20                                    & 75.57                                  & 67.25                                  & 69.19                                  & 65.91                                  & 39.08                                  & 40.80                                  \\
                                       & SliceGPT  \cite{ashkboosslicegpt}                                 & 7.10                                  & 50.58                                  & 35.49                                    & 65.18                                  & 65.67                                  & 52.30                                  & 59.26                                  & 36.77                                  & 39.40                                  \\
                                       & SVD-LLM  \cite{wang2024svd}                                    & 7.37                                  & 55.86                                  & 35.54                                    & 72.91                                  & 67.17                                  & 63.47                                  & 71.00                                  & 39.93                                  & 41.00                                  \\
\multirow{-6}{*}{\rotatebox{90}{\textbf{LLaMA-2 13B}}} & \cellcolor[HTML]{DAE8FC}\textbf{FLAT-LLM (ours)} & \cellcolor[HTML]{DAE8FC}\textbf{5.55} & \cellcolor[HTML]{DAE8FC}\textbf{63.00} & \cellcolor[HTML]{DAE8FC}\textbf{54.72}   & \cellcolor[HTML]{DAE8FC}75.84          & \cellcolor[HTML]{DAE8FC}\textbf{72.06} & \cellcolor[HTML]{DAE8FC}\textbf{73.36} & \cellcolor[HTML]{DAE8FC}\textbf{75.59} & \cellcolor[HTML]{DAE8FC}\textbf{46.25} & \cellcolor[HTML]{DAE8FC}43.20 \\ \hline \hline
                                       & Original                                     & 3.12                                  & 71.38                                  & 68.80                                    & 82.75                                  & 77.82                                  & 83.80                                  & 80.72                                  & 57.17                                  & 48.60                                  \\\cline{2-11}
                                       & FLAP  \cite{an2024fluctuation}                                     & 8.76                                  & 48.29                                  & 25.90                                    & 72.31                                  & 64.09                                  & 55.94                                  & 51.05                                  & 31.91                                  & 36.80                                  \\
                                       & SliceGPT  \cite{ashkboosslicegpt}                                 & 5.76                                  & 57.40                                  & 48.30                                    & 68.01                                  & 72.14                                  & 57.16                                  & 68.64                                  & 43.94                                  & 43.60                                  \\
                                       & SVD-LLM   \cite{wang2024svd}                                   & 5.96                                  & 61.07                                  & 52.10                                    & 74.48                                  & 72.61                                  & 68.41                                  & 71.93                                  & 46.93                                  & 41.00                                  \\
\multirow{-5}{*}{\rotatebox{90}{\textbf{LLaMA-2 70B}}} & \cellcolor[HTML]{DAE8FC}\textbf{FLAT-LLM (ours)} & \cellcolor[HTML]{DAE8FC}\textbf{4.33} & \cellcolor[HTML]{DAE8FC}\textbf{67.98} & \cellcolor[HTML]{DAE8FC}\textbf{67.35}   & \cellcolor[HTML]{DAE8FC}\textbf{77.20} & \cellcolor[HTML]{DAE8FC}\textbf{77.03} & \cellcolor[HTML]{DAE8FC}\textbf{78.44} & \cellcolor[HTML]{DAE8FC}\textbf{78.87} & \cellcolor[HTML]{DAE8FC}\textbf{51.19} & \cellcolor[HTML]{DAE8FC}\textbf{45.80} \\ \hline \hline
                                       & Original                                     & 4.92                                  & 68.14                                  & 62.50                                    & 82.05                                  & 73.95                                  & 81.02                                  & 79.55                                  & 53.92                                  & 44.00                                  \\\cline{2-11}
                                       & FLAP  \cite{an2024fluctuation}                                     & 7.11                                  & 48.29                                  & 25.90                                    & \textbf{72.31}                         & 64.09                                  & 55.94                                  & 51.05                                  & 31.91                                  & 36.80                                  \\
                                       & SliceGPT     \cite{ashkboosslicegpt}                              & 9.06                                  & 43.18                                  & 25.52                                    & 59.35                                  & 61.21                                  & 45.11                                  & 51.60                                  & 30.29                                  & 29.20                                  \\
                                       & SVD-LLM    \cite{wang2024svd}                                  & 9.29                                  & 50.23                                  & 25.02                                    & 70.46                                  & 64.56                                  & 58.09                                  & \textbf{69.28}                         & 37.63                                  & 26.60                                  \\
\multirow{-5}{*}{\rotatebox{90}{\textbf{Mistral-7B}}}  & \cellcolor[HTML]{DAE8FC}\textbf{FLAT-LLM (ours)} & \cellcolor[HTML]{DAE8FC}\textbf{6.11} & \cellcolor[HTML]{DAE8FC}\textbf{58.54} & \cellcolor[HTML]{DAE8FC}\textbf{57.21}   & \cellcolor[HTML]{DAE8FC}70.84          & \cellcolor[HTML]{DAE8FC}\textbf{69.30} & \cellcolor[HTML]{DAE8FC}\textbf{63.76} & \cellcolor[HTML]{DAE8FC}68.73          & \cellcolor[HTML]{DAE8FC}\textbf{41.72} & \cellcolor[HTML]{DAE8FC}\textbf{38.20} \\ \hline \hline
& Original                                     &  5.75                                  &  68.14                                 &  65.43                                   &  80.85                                & 73.40                                   & 79.17                                  &  80.09                                & 53.24                                  & 44.80                                  \\\cline{2-11}
& FLAP  \cite{an2024fluctuation}                                     &   8.42                                &  54.42                                 &  42.24                                   &  73.50                        &  65.90                                 & 59.87                                  & 64.22                               & 36.26                                  &   39.00                                  \\
& SliceGPT  \cite{ashkboosslicegpt}                                 &  16.62                                &  41.28                                  &  25.07                                   &    60.23                      &  57.22                                 &  40.46                                 &  47.26                                & 27.73                                  &   31.00                                \\
& SVD-LLM    \cite{wang2024svd}                                  & 17.17                                  &  47.06                                 &   28.64                                  &   66.27                       &  61.01                                 &  52.65                                 &  55.43                                & 31.66                                  &  33.80                                 \\
& \cellcolor[HTML]{DAE8FC}\textbf{FLAT-LLM (ours)} & \cellcolor[HTML]{DAE8FC}8.15 & \cellcolor[HTML]{DAE8FC}61.47         & \cellcolor[HTML]{DAE8FC}62.56 & \cellcolor[HTML]{DAE8FC}72.80 & \cellcolor[HTML]{DAE8FC}72.53 & \cellcolor[HTML]{DAE8FC}69.13          & \cellcolor[HTML]{DAE8FC}68.90 & \cellcolor[HTML]{DAE8FC}42.58  & \cellcolor[HTML]{DAE8FC}\textbf{41.80} \\
\multirow{-6}{*}{\rotatebox{90}{\textbf{LLaMA-2 7B}}}  & \cellcolor[HTML]{DAE8FC}\textbf{FLAT-LLM$^{\dag}$ (ours)} & \cellcolor[HTML]{DAE8FC}\textbf{8.02} & \cellcolor[HTML]{DAE8FC}\textbf{62.45} & \cellcolor[HTML]{DAE8FC}\textbf{63.21}   & \cellcolor[HTML]{DAE8FC}\textbf{73.67}          & \cellcolor[HTML]{DAE8FC}\textbf{74.11} & \cellcolor[HTML]{DAE8FC}\textbf{70.46}          & \cellcolor[HTML]{DAE8FC}\textbf{70.24} & \cellcolor[HTML]{DAE8FC}\textbf{44.28}          & \cellcolor[HTML]{DAE8FC}41.20 \\ \hline 
\end{tabular}
}
\label{table:acc20}
\end{table*}
\section{Experiments}
In this section, we first compare the performance of FLAT-LLM on language modeling and downstream tasks with recent fine-grained structural pruning methods across multiple LLM architectures and compression ratios. We then evaluate inference speedup and memory saving, and conduct ablation studies on our IPRS algorithm and the impact of calibration datasets. Additional experimental results, including calibration efficiency, performance under varying compression ratios on downstream tasks, and comparisons with coarse-grained importance-based pruning baselines, are provided in Appendix~\ref{appendix:addition}.
\subsection{Setups}
\paragraph{Models and Datasets.} We evaluate our method on multiple decoder-based generative models, including Llama-2 7B, 13B, 70B, Llama-3 8B \cite{touvron2023llama}, and Mistral 7B-v0.1 \cite{jiang2024mistral}.
Following the settings in previous works \cite{ashkboosslicegpt, wang2024svd}, we use 256 samples with 4096 tokens from WikiText-2 or Alpaca datasets for calibration. 
For the downstream evaluation, we use the LM Evaluation Harness \cite{eval-harness} and test on ten tasks: ARC-e, ARC-c, PIQA, WinoGrande, HellaSwag, BoolQ, OBQA, MathQA, CommonsenseQA, MMLU. Here, MMLU uses 5-shot evaluation, and all others use zero-shot.

\paragraph{Baselines.} We compare our method with recent decomposition-based pruning approaches, including SVD-LLM \cite{wang2024svd}, SliceGPT \cite{ashkboosslicegpt}, as well as fine-grained importance-based methods such as FLAP \cite{an2024fluctuation} and LLM-Pruner \cite{ma2023llm}. 
Additionally, we evaluate our approach against coarse-grained structural pruning techniques, including Pivoting Factorization \cite{zhaopivoting}, LLM Surgeon \cite{vanllm}, ShortGPT \cite{men2024shortgpt}, LaCo \cite{yang2024laco}, and BlockPruner \cite{zhong2024blockpruner}, as reported in Appendix~\ref{appendix:ppl} and ~\ref{appendix:acc}.

\paragraph{Implementation Details.} Our implementation is built on the Huggingface Transformers library \cite{wolf2019huggingface}. 
The MHA blocks are compressed using our proposed head-wise PCA, while the MLP modules are compressed using selection matrices derived from the Nyström approximation; implementation details are provided in Appendix~\ref{appendix:MLP}. 
Unless otherwise specified, all main-text experimental results are obtained using FLAT-LLM with compression applied only to the value and output projections in the attention module. We denote the variant that additionally compresses the query and key projections as FLAT-LLM$^\dag$.
All decomposition computations are performed in double precision to ensure numerical stability. Experiments are conducted on a single A100 40GB GPU, except for LLaMA-2 70B, which is evaluated using 4 A100 GPUs. All methods are evaluated without any recovery fine-tuning. 


\subsection{Performance Comparison}
\textbf{Performance on Different LLMs.} Table~\ref{table:acc20} compares the language modeling perplexity and downstream task accuracy of FLAT-LLM and FLAT-LLM$^\dag$ against prior structural compression methods across five LLMs under a 20\% compression ratio.
FLAT-LLM$^\dag$ augments FLAT-LLM by additionally applying head-wise compression to the query and key projections, which consistently improves accuracy under the same compression budget.
Across all evaluated models, FLAT-LLM($^\dag$) achieves the best overall performance in both average accuracy and perplexity, indicating strong generalization and text generation quality under compression.
For larger models, like LLaMA-2 13B and 70B, FLAT-LLM incurs only a modest average accuracy drop of 2.7\% and 3.4\%. In contrast to prior methods, which often struggle on newer architectures like Mistral-7B and LLaMA-3 8B, FLAT-LLM consistently maintains high performance, yielding up to 15.4\% and 20.2\% accuracy improvements, respectively.
Notably, FLAT-LLM outperforms all baselines by as much as 40\% on the MMLU 5-shot benchmark, demonstrating its effectiveness at preserving factual reasoning and broad-domain knowledge.

\textcolor{black}{We further evaluate zero-shot downstream performance under varying compression ratios (see Figure~\ref{fig:acc_model} in Appendix~\ref{appendix:acc}).
FLAT-LLM consistently outperforms prior low-rank decomposition baselines across all models and compression levels, demonstrating strong robustness and effectiveness.}

\begin{table}[htbp]
\caption{Perplexity comparison with GPTQ for LLaMA-2 7B and Llama-3 8B on WikiText-2.}
\resizebox{.45\textwidth}{!}{
\begin{tabular}{l|l|l|l}
\hline
\textbf{Model}                         & \textbf{Method}                                                                                   & \textbf{Size (GB)}          & \textbf{PPL} $\downarrow$                           \\ \hline
                                       & Original                                                                                          & 14                          & 5.12                                   \\\cline{2-4} 
                                       & GPTQ 2-bit                                                                                     & 1.8 (7.8 $\times$)                        & NaN                                    \\
\multirow{-3}{*}{\textbf{LLaMA-2 7B}}  & \cellcolor[HTML]{DAE8FC}\textbf{\begin{tabular}[c]{@{}l@{}}FLAT-LLM \\ + GPTQ 3-bit\end{tabular}} & \cellcolor[HTML]{DAE8FC}\textbf{1.8 (7.8 $\times$)  } & \cellcolor[HTML]{DAE8FC}\textbf{13.43} \\ \hline
                                       & Original                                                                                          & 16                          & 5.75                                   \\\cline{2-4} 
                                       & GPTQ 3-bit                                                                                       & 3.0 (5.3 $\times$)                        & 39.85                                  \\
\multirow{-3}{*}{\textbf{LLaMA-3 8B}} & \cellcolor[HTML]{DAE8FC}\textbf{\begin{tabular}[c]{@{}l@{}}FLAT-LLM \\ + GPTQ 4-bit\end{tabular}} & \cellcolor[HTML]{DAE8FC}\textbf{2.8 (5.7 $\times$)  } & \cellcolor[HTML]{DAE8FC}\textbf{11.91}  \\ \hline
\end{tabular}}
\label{table:quant}
\end{table}
\subsection{Comparison with Quantization}
\textcolor{black}{We further integrate FLAT-LLM with post-training quantization methods to assess their combined effectiveness. As shown in Table~\ref{table:quant}, applying FLAT-LLM (with a 30\% compression ratio) followed by GPTQ-3/4-bit quantization achieves lower perplexity than GPTQ-2/3-bit alone, while maintaining the same or even lower memory footprint.
This integration makes it possible to compress large language models by $5-8\times$ with negligible accuracy degradation, enabling highly efficient and effective model deployment.}

\begin{figure}
    \centering
    \includegraphics[width=\linewidth]{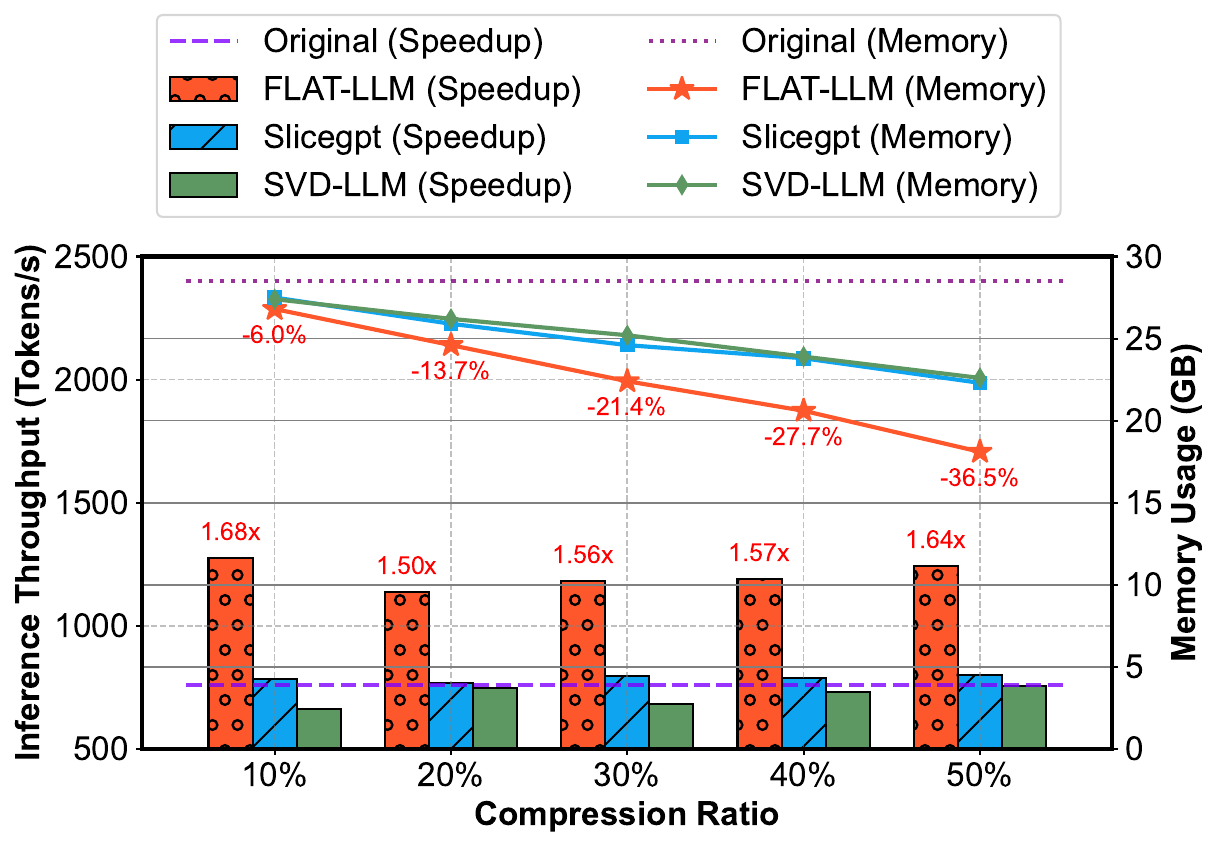}
    \caption{Comparison of inference throughput and memory usage with prior low-rank-based methods.}
    \label{fig:tops}
\end{figure}

\subsection{Inference Efficiency}
Figure~\ref{fig:tops} compares the inference throughput \textcolor{black}{and memory usage} of our method with prior low-rank approaches, including SliceGPT and SVD-LLM, across compression ratios ranging from 10\% to 50\% on LLaMA-2 7B. Following previous setups \cite{ashkboosslicegpt}, we generate 256 tokens with a batch size of 64, where we report the throughput during decoding and the CUDA memory usage after decoding.
Without any CUDA-level optimization, our method consistently outperforms the original model, achieving speedups over 1.50$\times$ across all compression ratios.
Notably, even at a modest 10\% overall compression ratio, our method attains a 1.68$\times$ throughput improvement, as the proposed non-uniform rank selection strategy concentrates compression on middle-to-late layers, which are empirically more memory-bound in inference.
\textcolor{black}{Additionally, due to the reduced dimensionality of the value outputs, FLAT-LLM further reduces activation memory in the value cache. As a result, even under the same compression ratio, it achieves the lowest memory usage among all low-rank model compression methods.}
Compared to SliceGPT and SVD-LLM, it achieves up to $1.63\times$ and $1.93\times$ higher throughput, and reduces memory usage by $19\%$ and $20\%$, respectively. 
These results highlight the efficiency of FLAT-LLM, making it a strong candidate for real-world deployment.

\subsection{Ablation Study}
\begin{figure}
    \centering
    \includegraphics[width=.9\linewidth]{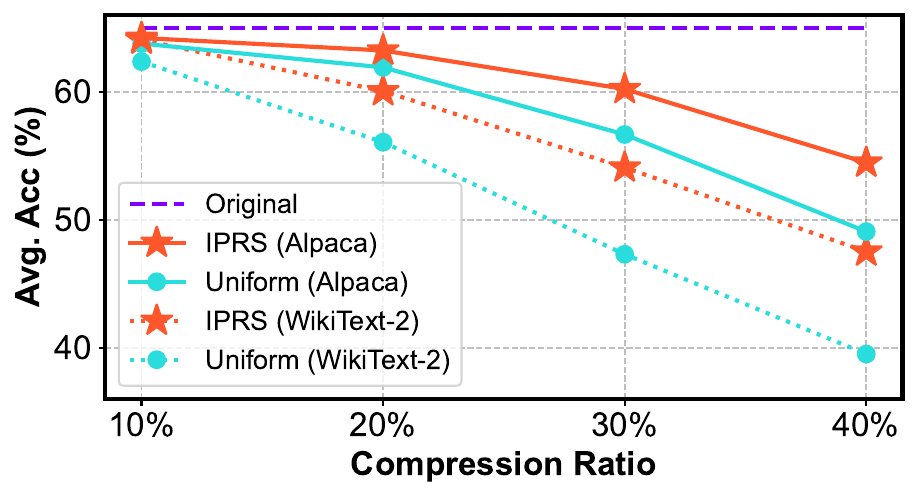}
    \caption{Comparison of zero-shot average accuracy on downstream datasets versus compression ratio using uniform rank and our IPRS algorithm on Llama-2 13B.}
    \label{fig:Abla}
\end{figure}

In this section, we present an ablation study of our rank selection method and evaluate its performance using different calibration datasets. 
Specifically, we report the average accuracy across eight downstream tasks using calibration performed on either the WikiText-2 or Alpaca dataset.
\paragraph{Importance-Preserving Rank Selection.}
To evaluate the effectiveness of our Importance-Preserving Rank Selection (IPRS) method, Figure~\ref{fig:Abla} compares the average zero-shot precision of different rank allocation strategies - uniform and IPRS - between compression ratios ranging from 10\% to 40\% on LLaMA-2 13B.
As shown in Figure~\ref{fig:Abla}, IPRS consistently outperforms the uniform baseline across all compression settings. The performance gap becomes more pronounced at higher compression ratios, achieving gains of $5.4\%$ with Alpaca calibration and $8.0\%$ with WikiText-2 at a 40\% compression ratio. This is because sensitive layers suffer disproportionately high truncation loss under aggressive compression, making adaptive rank selection increasingly important. These results underscore the complementary strengths of fine-grained PCA and importance-aware rank allocation. In Appendix~\ref{appendix:IPRS2}, we further apply IPRS to the SVD-LLM method, which also shows improved performance over uniform rank.  
\textcolor{black}{In Appendix~\ref{appendix:modegpt}, we additionally show that IPRS consistently outperforms MoDeGPT’s rank allocation strategy.}

\begin{figure}
    \centering
    \includegraphics[width=\linewidth]{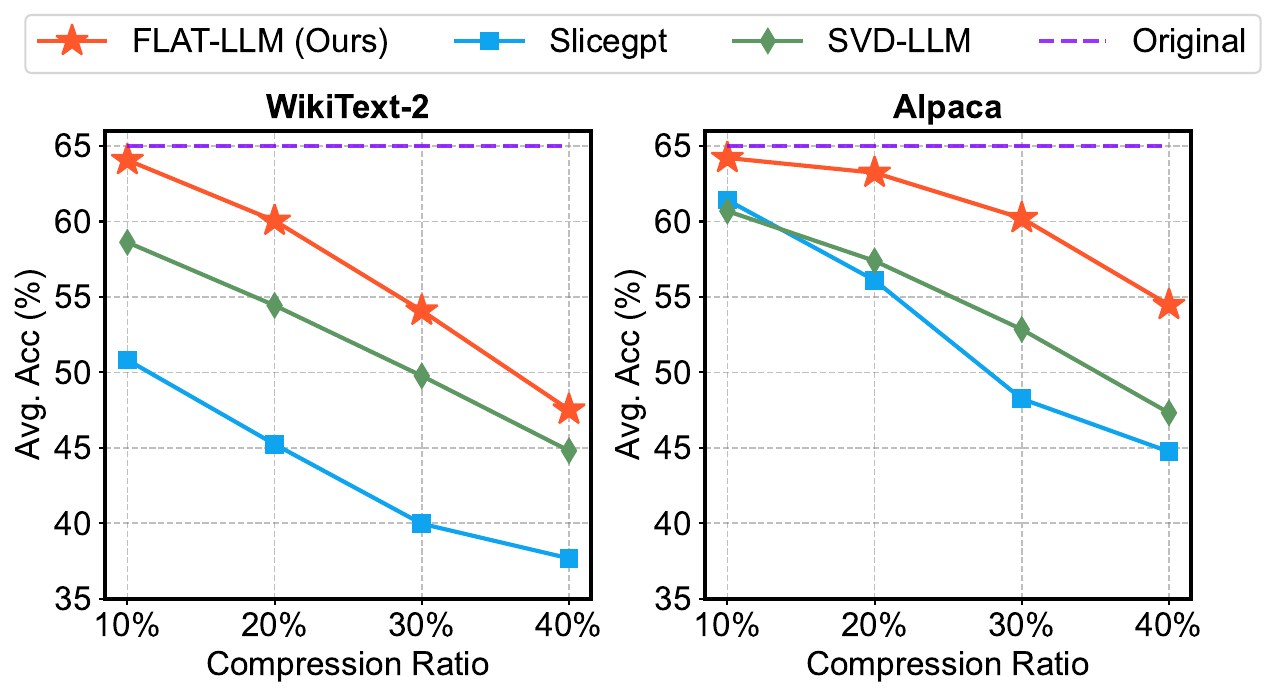}
    \caption{Comparison of zero-shot average accuracy on downstream datasets versus compression ratio calibrated on WikiText-2 or Alpaca on Llama-2 13B.}
    \label{fig:Calidata}
\end{figure}
\paragraph{Calibration Dataset.}
We evaluate the average zero-shot accuracy on LLaMA-2 13B using two calibration datasets—WikiText-2 and Alpaca. As shown in Figure~\ref{fig:Calidata}, FLAT-LLM consistently outperforms SliceGPT and SVD-LLM across all settings and compression levels, demonstrating strong generalization with WikiText-2 calibration and even greater gains with Alpaca. Notably, under Alpaca calibration, FLAT-LLM maintains high accuracy, with less than a 5\% drop observed at up to 30\% compression. These results highlight the robustness of FLAT-LLM across diverse tasks, calibration datasets, and compression regimes.

\section{Conclusion}
We propose a fast and accurate training-free structural compression method for LLMs, leveraging low-rank transformations in the activation space. By combining an importance-preserving global rank allocation strategy with efficient head-wise PCA-based approximations, our approach delivers model with strong generation performance using minimal calibration time. Experiments on LLaMA and Mistral show superior generalization and substantial inference speedups compared to prior structural pruning and decomposition-based methods.

\section*{Limitations}
While FLAT-LLM demonstrates strong empirical performance in compressing large language models without fine-tuning, several limitations remain. 
First, although FLAT-LLM achieves superior performance and reduced memory consumption when combined with post-training quantization methods such as GPTQ, further CUDA kernel–level optimization is required to fully realize the potential inference speedup in practice.
Second, while FLAT-LLM attains substantial throughput acceleration even at a 10\% compression ratio due to its non-uniform rank selection strategy, we do not explicitly analyze how the resulting rank distribution influences end-to-end speedup, nor do we design a hardware-aware rank allocation scheme.
We leave both directions for future work.

\section*{Acknowledgments}
This work is co-funded by Intel Strategic Research Sectors (SRS) - Systems Integration SRS \& Devices SRS.

\bibliography{anthology}

@article{touvron2023llama,
  title={Llama: Open and efficient foundation language models},
  author={Touvron, Hugo and Lavril, Thibaut and Izacard, Gautier and Martinet, Xavier and Lachaux, Marie-Anne and Lacroix, Timoth{\'e}e and Rozi{\`e}re, Baptiste and Goyal, Naman and Hambro, Eric and Azhar, Faisal and others},
  journal={arXiv preprint arXiv:2302.13971},
  year={2023}
}

@article{bai2023qwen,
  title={Qwen technical report},
  author={Bai, Jinze and Bai, Shuai and Chu, Yunfei and Cui, Zeyu and Dang, Kai and Deng, Xiaodong and Fan, Yang and Ge, Wenbin and Han, Yu and Huang, Fei and others},
  journal={arXiv preprint arXiv:2309.16609},
  year={2023}
}

@article{liu2024deepseek,
  title={Deepseek-v3 technical report},
  author={Liu, Aixin and Feng, Bei and Xue, Bing and Wang, Bingxuan and Wu, Bochao and Lu, Chengda and Zhao, Chenggang and Deng, Chengqi and Zhang, Chenyu and Ruan, Chong and others},
  journal={arXiv preprint arXiv:2412.19437},
  year={2024}
}

@inproceedings{huang2024billm,
  title={BiLLM: pushing the limit of post-training quantization for LLMs},
  author={Huang, Wei and Liu, Yangdong and Qin, Haotong and Li, Ying and Zhang, Shiming and Liu, Xianglong and Magno, Michele and Qi, Xiaojuan},
  booktitle={Proceedings of the 41st International Conference on Machine Learning},
  pages={20023--20042},
  year={2024}
}

@inproceedings{tian2023bebert,
  title={BEBERT: Efficient and robust binary ensemble BERT},
  author={Tian, Jiayi and Fang, Chao and Wang, Haonan and Wang, Zhongfeng},
  booktitle={ICASSP 2023-2023 IEEE International Conference on Acoustics, Speech and Signal Processing (ICASSP)},
  pages={1--5},
  year={2023},
  organization={IEEE}
}

@inproceedings{jiao2020tinybert,
  title={TinyBERT: Distilling BERT for Natural Language Understanding},
  author={Jiao, Xiaoqi and Yin, Yichun and Shang, Lifeng and Jiang, Xin and Chen, Xiao and Li, Linlin and Wang, Fang and Liu, Qun},
  booktitle={Findings of the Association for Computational Linguistics: EMNLP 2020},
  pages={4163--4174},
  year={2020}
}

@inproceedings{sun2020mobilebert,
  title={MobileBERT: a Compact Task-Agnostic BERT for Resource-Limited Devices},
  author={Sun, Zhiqing and Yu, Hongkun and Song, Xiaodan and Liu, Renjie and Yang, Yiming and Zhou, Denny},
  booktitle={Proceedings of the 58th Annual Meeting of the Association for Computational Linguistics},
  pages={2158--2170},
  year={2020}
}

@article{yang2025wanda,
  title={Wanda++: Pruning Large Language Models via Regional Gradients},
  author={Yang, Yifan and Zhen, Kai and Ganesh, Bhavana and Galstyan, Aram and Huybrechts, Goeric and M{\"u}ller, Markus and K{\"u}bler, Jonas M and Swaminathan, Rupak Vignesh and Mouchtaris, Athanasios and Bodapati, Sravan Babu and others},
  journal={arXiv preprint arXiv:2503.04992},
  year={2025}
}

@inproceedings{sunsimple,
  title={A Simple and Effective Pruning Approach for Large Language Models},
  author={Sun, Mingjie and Liu, Zhuang and Bair, Anna and Kolter, J Zico},
  booktitle={The Twelfth International Conference on Learning Representations},
  year={2024}
}

@inproceedings{vanllm,
  title={The LLM Surgeon},
  author={van der Ouderaa, Tycho FA and Nagel, Markus and Van Baalen, Mart and Blankevoort, Tijmen},
  booktitle={The Twelfth International Conference on Learning Representations},
  year={2024}
}

@article{men2024shortgpt,
  title={Shortgpt: Layers in large language models are more redundant than you expect},
  author={Men, Xin and Xu, Mingyu and Zhang, Qingyu and Wang, Bingning and Lin, Hongyu and Lu, Yaojie and Han, Xianpei and Chen, Weipeng},
  journal={arXiv preprint arXiv:2403.03853},
  year={2024}
}

@inproceedings{yang2024loretta,
  title={LoRETTA: Low-Rank Economic Tensor-Train Adaptation for Ultra-Low-Parameter Fine-Tuning of Large Language Models},
  author={Yang, Yifan and Zhou, Jiajun and Wong, Ngai and Zhang, Zheng},
  booktitle={Proceedings of the 2024 Conference of the North American Chapter of the Association for Computational Linguistics: Human Language Technologies (Volume 1: Long Papers)},
  pages={3161--3176},
  year={2024}
}

@inproceedings{yu2023compressing,
  title={Compressing transformers: features are low-rank, but weights are not!},
  author={Yu, Hao and Wu, Jianxin},
  booktitle={Proceedings of the AAAI Conference on Artificial Intelligence},
  volume={37},
  number={9},
  pages={11007--11015},
  year={2023}
}

@inproceedings{gao2024adaptive,
  title={Adaptive rank selections for low-rank approximation of language models},
  author={Gao, Shangqian and Hua, Ting and Hsu, Yen-Chang and Shen, Yilin and Jin, Hongxia},
  booktitle={Proceedings of the 2024 Conference of the North American Chapter of the Association for Computational Linguistics: Human Language Technologies (Volume 1: Long Papers)},
  pages={227--241},
  year={2024}
}

@inproceedings{ashkboosslicegpt,
  title={SliceGPT: Compress Large Language Models by Deleting Rows and Columns},
  author={Ashkboos, Saleh and Croci, Maximilian L and do Nascimento, Marcelo Gennari and Hoefler, Torsten and Hensman, James},
  booktitle={The Twelfth International Conference on Learning Representations},
  year={2024}
}

@article{lin2024modegpt,
  title={Modegpt: Modular decomposition for large language model compression},
  author={Lin, Chi-Heng and Gao, Shangqian and Smith, James Seale and Patel, Abhishek and Tuli, Shikhar and Shen, Yilin and Jin, Hongxia and Hsu, Yen-Chang},
  journal={arXiv preprint arXiv:2408.09632},
  year={2024}
}

@article{ma2023llm,
  title={Llm-pruner: On the structural pruning of large language models},
  author={Ma, Xinyin and Fang, Gongfan and Wang, Xinchao},
  journal={Advances in neural information processing systems},
  volume={36},
  pages={21702--21720},
  year={2023}
}

@article{yuan2023asvd,
  title={Asvd: Activation-aware singular value decomposition for compressing large language models},
  author={Yuan, Zhihang and Shang, Yuzhang and Song, Yue and Wu, Qiang and Yan, Yan and Sun, Guangyu},
  journal={arXiv preprint arXiv:2312.05821},
  year={2023}
}

@article{wang2024svd,
  title={Svd-llm: Truncation-aware singular value decomposition for large language model compression},
  author={Wang, Xin and Zheng, Yu and Wan, Zhongwei and Zhang, Mi},
  journal={arXiv preprint arXiv:2403.07378},
  year={2024}
}

@misc{eval-harness,
  author       = {Gao, Leo and Tow, Jonathan and Abbasi, Baber and Biderman, Stella and Black, Sid and DiPofi, Anthony and Foster, Charles and Golding, Laurence and Hsu, Jeffrey and Le Noac'h, Alain and Li, Haonan and McDonell, Kyle and Muennighoff, Niklas and Ociepa, Chris and Phang, Jason and Reynolds, Laria and Schoelkopf, Hailey and Skowron, Aviya and Sutawika, Lintang and Tang, Eric and Thite, Anish and Wang, Ben and Wang, Kevin and Zou, Andy},
  title        = {A framework for few-shot language model evaluation},
  month        = 07,
  year         = 2024,
  publisher    = {Zenodo},
  version      = {v0.4.3},
  doi          = {10.5281/zenodo.12608602},
  url          = {https://zenodo.org/records/12608602}
}

@misc{jiang2024mistral,
      title={Mistral 7B}, 
      author={Albert Q. Jiang and Alexandre Sablayrolles and Arthur Mensch and Chris Bamford and Devendra Singh Chaplot and Diego de las Casas and Florian Bressand and Gianna Lengyel and Guillaume Lample and Lucile Saulnier and Lélio Renard Lavaud and Marie-Anne Lachaux and Pierre Stock and Teven Le Scao and Thibaut Lavril and Thomas Wang and Timothée Lacroix and William El Sayed},
      year={2023},
      eprint={2310.06825},
      archivePrefix={arXiv},
      primaryClass={cs.CL},
}

@inproceedings{ainslie2023gqa,
  title={GQA: Training Generalized Multi-Query Transformer Models from Multi-Head Checkpoints},
  author={Ainslie, Joshua and Lee-Thorp, James and de Jong, Michiel and Zemlyanskiy, Yury and Lebron, Federico and Sanghai, Sumit},
  booktitle={Proceedings of the 2023 Conference on Empirical Methods in Natural Language Processing},
  pages={4895--4901},
  year={2023}
}

@article{zhong2024blockpruner,
  title={Blockpruner: Fine-grained pruning for large language models},
  author={Zhong, Longguang and Wan, Fanqi and Chen, Ruijun and Quan, Xiaojun and Li, Liangzhi},
  journal={arXiv preprint arXiv:2406.10594},
  year={2024}
}

@inproceedings{yang2024laco,
  title={LaCo: Large Language Model Pruning via Layer Collapse},
  author={Yang, Yifei and Cao, Zouying and Zhao, Hai},
  booktitle={Findings of the Association for Computational Linguistics: EMNLP 2024},
  pages={6401--6417},
  year={2024}
}

@inproceedings{an2024fluctuation,
  title={Fluctuation-based adaptive structured pruning for large language models},
  author={An, Yongqi and Zhao, Xu and Yu, Tao and Tang, Ming and Wang, Jinqiao},
  booktitle={Proceedings of the AAAI Conference on Artificial Intelligence},
  volume={38},
  number={10},
  pages={10865--10873},
  year={2024}
}

@article{wolf2019huggingface,
  title={Huggingface's transformers: State-of-the-art natural language processing},
  author={Wolf, Thomas and Debut, Lysandre and Sanh, Victor and Chaumond, Julien and Delangue, Clement and Moi, Anthony and Cistac, Pierric and Rault, Tim and Louf, R{\'e}mi and Funtowicz, Morgan and others},
  journal={arXiv preprint arXiv:1910.03771},
  year={2019}
}

@inproceedings{zhaopivoting,
  title={Pivoting Factorization: A Compact Meta Low-Rank Representation of Sparsity for Efficient Inference in Large Language Models},
  author={Zhao, Jialin and Zhang, Yingtao and Cannistraci, Carlo Vittorio},
  booktitle={Forty-second International Conference on Machine Learning},
  year={2025}
}
\clearpage
\appendix
\section{Head-wise PCA for Grouped Query Attention}
\label{appendix:GQA}
\paragraph{Grouped Query Attention.} 
Modern architectures such as Llama-3 and Mistral employ Grouped-Query Attention (GQA), in which multiple query heads share a smaller set of key-value heads. 
This design aims to improve the inference efficiency by reducing both the memory footprint and computation cost associated with the KV cache.

Let \( H \) and \( G \) denote the number of query and key-value heads, where typically \( H > G \). 
For each query head \( h \in \{1, \dots, H\} \), let its associated key-value head be denoted by \( g(h) \in \{1, \dots, G\}\). 
In most cases, where \( H = nG \) for some integer \( n \), the mapping is defined as \( g(h) = \left\lfloor \frac{h}{n} \right\rfloor \).
Under this setting, we reformulate Equation~\eqref{eqa:1} as follows:
\begin{align}
\begin{split}
\mathbf{A}^h &= \frac{ \mathbf{X}{\mathbf{W}_q^h}{^\top} (\mathbf{X}{\mathbf{W}_k^{g(h)}}{^\top}){^\top} }{ \sqrt{d_h} }, \\
\mathbf{Y}_v^h &= \text{Softmax}(\mathbf{A}^h)\mathbf{X}{\mathbf{W}_v^{g(h)}}{^\top} , \\ 
\end{split} 
\end{align}
where the key difference lies in the use of shared key and value projection $\mathbf{W}_k^{g(h)}, \mathbf{W}_v^{g(h)}$ across multiple query heads. 
Similarly, Equation~\eqref{eqa:2} can be rewritten as: 
\begin{align}
\mathbf{Y}_o^h =  \text{Softmax}(\mathbf{A}^h)\mathbf{X} {\mathbf{W}_v^{g(h)}}{^\top}{\mathbf{W}_o^h}{^\top}.
\label{eqa:11}
\end{align}

\begin{table*}[htbp]
\centering
\caption{Comparison of zero-shot downstream performance with prior importance-based compression methods for Llama-2 7B at 20\% compression ratio and Llama-2 13B at 30\% compression ratio.}
\resizebox{\textwidth}{!}{
\begin{tabular}{l|c|c|ccccc|c|c}
\hline
\multicolumn{1}{c|}{\textbf{Model}}    & \textbf{Method}                                                  & \textbf{Ratio}              & \textbf{WinoG.}                    & \textbf{HellaS.}                     & \textbf{ARC-e}                         & \textbf{ARC-c}                         & \textbf{PIQA}                          & \textbf{Avg.}                          & \textbf{delta}                         \\ \hline
                                       & Original                                                         & 0\%                          & 69.06                                  & 75.99                                  & 74.58                                  & 46.25                                  & 77.91                                  & 68.76                                  & 0                                      \\ \cline{2-10} 
                                       & LaCo \cite{yang2024laco}                                                            &                             & 60.46                                  & 54.08                                  & 55.39                                  & 35.84                                  & 68.34                                  & 54.82                                  & -13.94                                 \\
                                       & ShortGPT \cite{men2024shortgpt}                                                        &                             & 65.90                                  & 62.63                                  & 56.06                                  & 36.09                                  & 70.24                                  & 58.18                                  & -10.58                                 \\
                                       & BlockPruner \cite{zhong2024blockpruner}                                                     & \multirow{-3}{*}{20\%}       & 62.43                                  & 65.87                                  & 61.07                                  & 37.29                                  & 74.21                                  & 60.17                                  & -8.59                                  \\
\multirow{-5}{*}{\textbf{LLaMA-2 7B}}  & \cellcolor[HTML]{DAE8FC}{\color[HTML]{333333} \textbf{FLAT-LLM (ours)}} & \cellcolor[HTML]{DAE8FC}20\% & \cellcolor[HTML]{DAE8FC}\textbf{67.88} & \cellcolor[HTML]{DAE8FC}\textbf{69.24} & \cellcolor[HTML]{DAE8FC}\textbf{70.45} & \cellcolor[HTML]{DAE8FC}\textbf{41.38} & \cellcolor[HTML]{DAE8FC}\textbf{75.35} & \cellcolor[HTML]{DAE8FC}\textbf{64.86} & \cellcolor[HTML]{DAE8FC}\textbf{-3.90} \\ \hline
                                       & Original                                                         & 0\%                          & 72.22                                  & 79.39                                  & 79.42                                  & 49.06                                  & 80.52                                  & 72.12                                  & 0                                      \\ \cline{2-10} 
                                       & LaCo \cite{yang2024laco}                                                            &                             & 59.27                                  & 60.44                                  & 54.34                                  & 34.56                                  & 72.42                                  & 55.44                                  & -16.68                                 \\
                                       & ShortGPT \cite{men2024shortgpt}                                                        &                             & 70.80                                  & 67.80                                  & 60.35                                  & 41.30                                  & 72.74                                  & 62.60                                  & -9.52                                  \\
                                       & BlockPruner \cite{zhong2024blockpruner}                                                     & \multirow{-3}{*}{25\%}       & 66.30                                  & 72.20                                  & 65.82                                  & 41.38                                  & 76.93                                  & 64.53                                  & -7.59                                  \\
\multirow{-5}{*}{\textbf{LLaMA-2 13B}} & \cellcolor[HTML]{DAE8FC}{\color[HTML]{333333} \textbf{FLAT-LLM (ours)}} & \cellcolor[HTML]{DAE8FC}25\% & \cellcolor[HTML]{DAE8FC}\textbf{71.35} & \cellcolor[HTML]{DAE8FC}\textbf{72.93} & \cellcolor[HTML]{DAE8FC}\textbf{73.48} & \cellcolor[HTML]{DAE8FC}\textbf{43.94} & \cellcolor[HTML]{DAE8FC}\textbf{76.44} & \cellcolor[HTML]{DAE8FC}\textbf{67.63} & \cellcolor[HTML]{DAE8FC}\textbf{-4.49} \\ \hline
\end{tabular}
}
\label{table:acc_alpaca}
\end{table*}

\paragraph{Head-wise PCA for GQA}
Although GQA introduces a mismatch between the number of heads in the value and output layers, the joint compression technique remains applicable to GQA-based architectures.

Let \( \mathbf{Y}_v^{g(h)} \) represent the output of the value layer for key-value head \( g(h) \). By applying PCA, we can transform the value output as 
$$\mathbf{Y}_v^{g(h)} = \mathbf{Y}_v^{g(h)} \mathbf{Q}_v^{g(h)} {\mathbf{Q}_v^{g(h)}}^{\top},$$
and reformulate Equation~\eqref{eqa:11} as
\begin{align}
&\mathbf{Y}_o^h =  \text{Softmax}(\mathbf{A}^h)\mathbf{X} {\mathbf{W}_v^{g(h)}}{^\top}{\mathbf{Q}_v^{g(h)}}{\mathbf{Q}_v^{g(h)}}{^\top}{\mathbf{W}_o^h}{^\top}. \nonumber
\end{align}

Even though the output layer contains \( H \) heads and the value layer contains only \( G \), each output projection head \( h \) uses the PCA basis \( \mathbf{Q}_v^{g(h)} \) derived from its corresponding value head \( g(h) \). This enables joint compression of both the value and output projection layers under GQA. The process can be expressed as:

\begin{align}
\begin{split}
&\tilde{\mathbf{Y}}_{o}^{h} = \text{Softmax}(\mathbf{A}^{h})\mathbf{X}{\mathbf{W}_v^{g(h)}}{^\top}\tilde{\mathbf{Q}}_v^{g(h)}{\tilde{\mathbf{Q}}_v^{g(h)}}{^\top}{\mathbf{W}_o^{h}}{^\top}, \nonumber \\
&\tilde{\mathbf{Y}}_{o}^{h} = \text{Softmax}(\mathbf{A}^{h})\mathbf{X}\tilde{\mathbf{W}}_v^{g(h)}{^\top}\tilde{\mathbf{W}}_o^{h}{^\top}, \nonumber
\end{split} 
\end{align}
where the first equation represents truncation and second denotes absoprtion. The PCA basis \( \tilde{\mathbf{Q}}_v^{g(h)} \in \mathbb{R}^{d_h \times r} \) is truncated to rank \( r \). As a result, the output layer shares \( G \) PCA basis, reducing the total computation required for PCA. After truncation, the shared value projection matrix and the per-head output projection matrix become \( \tilde{\mathbf{W}}_v^{g(h)} \in \mathbb{R}^{r \times d_{\text{hid}}} \) and \( \tilde{\mathbf{W}}_o^h \in \mathbb{R}^{d_{\text{hid}} \times r} \), respectively.

\section{Truncation Loss Analysis}
\subsection{Theoretical Proof}
\label{appendix:proof}
In the following, we provide the detailed proof on the direct mapping between eigenvalues and truncation loss in single-head and multi-head cases.


\noindent\textbf{Theorem 4.1} [Reconstruction Error of Single-head Output PCA Projection]
\textit{Let \( \mathbf{Y}_v^h =  \mathbf{X} {\mathbf{W}_v^h}{^\top}\), and \( \tilde{\mathbf{Y}}_v^h = \mathbf{Y}_v^h\tilde{\mathbf{Q}}_v^h {\tilde{\mathbf{Q}}_v^h}{^\top} \) be the rank-\( r \) approximation obtained by projecting \( \mathbf{Y}_v^h \) onto its top-\( r \) principal components. Then the squared Frobenius norm of the reconstruction error satisfies:
\vspace{-0.2cm}
\[\| \mathbf{Y}_v^h - \tilde{\mathbf{Y}}_v^h \|_F^2 = \sum_{i = r+1}^{d_h} \lambda_{i}^h,\]
\vspace{-0.2cm}
where \( \{ \lambda_i^h \} \) are the eigenvalues of \( {\mathbf{Y}^h_v}{^\top} \mathbf{Y}_v^h \).}

\begin{proof}
The projection \( \tilde{\mathbf{Y}}_v^h = \mathbf{Y}_v^h\tilde{\mathbf{Q}}_v^h {\tilde{\mathbf{Q}}_v^h}{^\top} \) minimizes the Frobenius norm among all rank-\( r \) approximations. Let \( \mathbf{P} = \tilde{\mathbf{Q}}_v^h {\tilde{\mathbf{Q}}_v^h}{^\top} \in \mathbb{R}^{d_h \times d_h} \) be the orthogonal projector onto the top-\( r \) eigenspace. The squared Frobenius norm of the reconstruction error is:
\begin{align}
\begin{split}
&\| \mathbf{Y}_v^h - \tilde{\mathbf{Y}}_v^h \|_F^2 = \| \mathbf{Y}_v^h(\mathbf{I} - \mathbf{P})  \|_F^2 \\
&= \mathrm{Trace}\left[ {\mathbf{Y}_v^h} (\mathbf{I} - \mathbf{P}) (\mathbf{I} - \mathbf{P}){^\top} {\mathbf{Y}_v^h}{^\top} \right] \\
&= \mathrm{Trace}\left[ {\mathbf{Y}_v^h} (\mathbf{I} - \mathbf{P}) {\mathbf{Y}_v^h}{^\top} \right] \\
&= \mathrm{Trace}\left[ (\mathbf{I} - \mathbf{P}) {\mathbf{Y}_v^h}{^\top}{\mathbf{Y}_v^h} \right]. \nonumber
\end{split}
\end{align}

Since \({\mathbf{Y}^h_v}{^\top}{\mathbf{Y}_v^h}  = \mathbf{Q}_v^h \boldsymbol{\Lambda}_v^h {\mathbf{Q}_v^h}{^\top} \), let \(\hat{\mathbf{Q}_v^h} = \mathbf{Q}_v^h-\tilde{\mathbf{Q}}_v^h \), we have \( \mathbf{I} - \mathbf{P} =\hat{\mathbf{Q}_v^h} {\hat{\mathbf{Q}_v^h}}{^\top}  \), and thus:
\begin{align}
\begin{split}
&\| \mathbf{Y}_v^h - \tilde{\mathbf{Y}}_v^h \|_F^2 = \mathrm{Trace}\left[ \hat{\mathbf{Q}_v^h} {\hat{\mathbf{Q}_v^h}}{^\top} \mathbf{Q}_v^h \boldsymbol{\Lambda}_v^h {\mathbf{Q}_v^h}{^\top} \right] \\
&= \mathrm{Trace}\left[ {\hat{\mathbf{Q}_v^h}}{^\top} \mathbf{Q}_v^h \boldsymbol{\Lambda}_v^h {\mathbf{Q}_v^h}{^\top} \hat{\mathbf{Q}_v^h} \right] \\
&= \mathrm{Trace}\left[ \hat{\boldsymbol{\Lambda}_v^h} \right] = \sum_{i=r+1}^{d_h} \lambda_i^h. \nonumber
\end{split}
\end{align}
\end{proof}

\noindent\textbf{Corollary 4.2} [Reconstruction Error of Multi-head Output PCA Projection]
\textit{Let $\mathbf{Y}_v = \text{concat}(\mathbf{Y}_v^1,...\mathbf{Y}_v^H)$ be the concatenated output hidden states. 
The squared Frobenius norm of the reconstruction error satisfies:
\[
\| \mathbf{Y}_v - \tilde{\mathbf{Y}}_v \|_F^2 = \sum_{h=1}^H\sum_{i = r+1}^{d_h} \lambda_i^h,
\]}

\begin{proof}
\begin{align}
\begin{split}
&\| \mathbf{Y}_v - \tilde{\mathbf{Y}}_v \|_F^2 =\| \text{concat}(\mathbf{Y}_v^h - \tilde{\mathbf{Y}_v^h}) \|_F^2 \\
&= \sum_{h=1}^H\| (\mathbf{Y}_v^h - \tilde{\mathbf{Y}_v^h}) \|_F^2 = \sum_{h=1}^H \sum_{i=r+1}^{d_h} \lambda_i^h. \nonumber
\end{split}
\end{align}

\end{proof}


Therefore, the reconstruction loss of multi-head value output equals to the sum of the dropped eigenvalues of all heads. Truncating the smallest $d_h-r$ eigenvalues of each head leads to the lowest reconstruction loss.

\subsection{Empirical Results}
\label{appendix:emprical_error}
\begin{table}[htbp]
\centering
\caption{Comparison of layer-wise reconstruction error for SVD-LLM and FLAT-LLM under 20\% sparsity.}
\resizebox{.5\textwidth}{!}{
\begin{tabular}{c|c|c|c}
\hline
\textbf{Method} & \textbf{Layer 0} & \textbf{Layer 10} & \textbf{Layer 20} \\ \hline
SVD-LLM \cite{wang2024svd} & 1.56 & 1.56 & 1.53 \\
\rowcolor[HTML]{DAE8FC} 
\textbf{FLAT-LLM (ours)} & \textbf{1.53} & \textbf{1.52} & \textbf{1.52} \\ \hline
\end{tabular}
}
\label{tab:error}
\end{table}
\textcolor{black}{To evaluate the reconstruction quality of the attention block, we conducted an empirical study measuring the relative Frobenius norm between the compressed and original attention outputs. For a fair comparison, we used the same value-layer inputs from the original model at each layer for both SVD-LLM and FLAT-LLM under a uniform rank setting, thereby avoiding the influence of inter-layer error propagation. We evaluated the reconstruction error at a 20\% compression ratio for both methods on LLaMA-2 7B. The results show that FLAT-LLM consistently yields lower reconstruction error, indicating that our head-wise PCA design for the attention block more effectively preserves the low-rank structure of the activation space than SVD-LLM.}

\section{Nyström-based Low-rank MLP Approximation}
\label{appendix:MLP}
To compress the MLP layers, we apply a structured Nyström approximation guided by data-dependent ridge leverage scores inspired by MoDeGPT \cite{lin2024modegpt}.
\textcolor{black}{We further improve its efficiency by eliminating the need for the time-consuming pseudo-inverse computation, reducing the overall runtime from hours in MoDeGPT to only a few minutes in our approach.}
Let $d_{{hid}}$ and $d_{\text{int}}$ denote the hidden and intermediate dimensions, respectively. 
The method operates on the up-projection matrix $\mathbf{W}_1 \in \mathbb{R}^{d_{hid} \times d_{int}}$ and the down-projection matrix $\mathbf{W}_2 \in \mathbb{R}^{d_{int} \times d_{hid}}$ within each MLP block. 
To capture activation structure, we compute the correlation matrix $C_\sigma$ over intermediate hidden states passed through the SiLU activation. 
Ridge leverage scores are then derived from the regularized correlation matrix via $C_\sigma(C_\sigma + \mathbf{I})^{-1}$, quantifying the relative importance of each intermediate channel. 
Based on these scores, we construct a selection matrix $\mathbf{S}_k \in \mathbb{R}^{d_{{int}} \times k}$ that retains the top-$k$ most informative channels, where $k$ is determined by the target sparsity ratio. The up-projection is compressed by selecting the top-$k$ columns, i.e., $\mathbf{W}_1 \mathbf{S}_k$, while the down-projection is approximated via Nyström reconstruction: $\left(\mathbf{S}_k^\top C_\sigma \mathbf{S}_k\right)^{-1} \mathbf{S}_k^\top C_\sigma \mathbf{W}_2$. 
This data-aware procedure preserves key activation subspaces while significantly reducing parameter count and computation.

\begin{figure*}
    \centering
    \includegraphics[width=.9\linewidth]{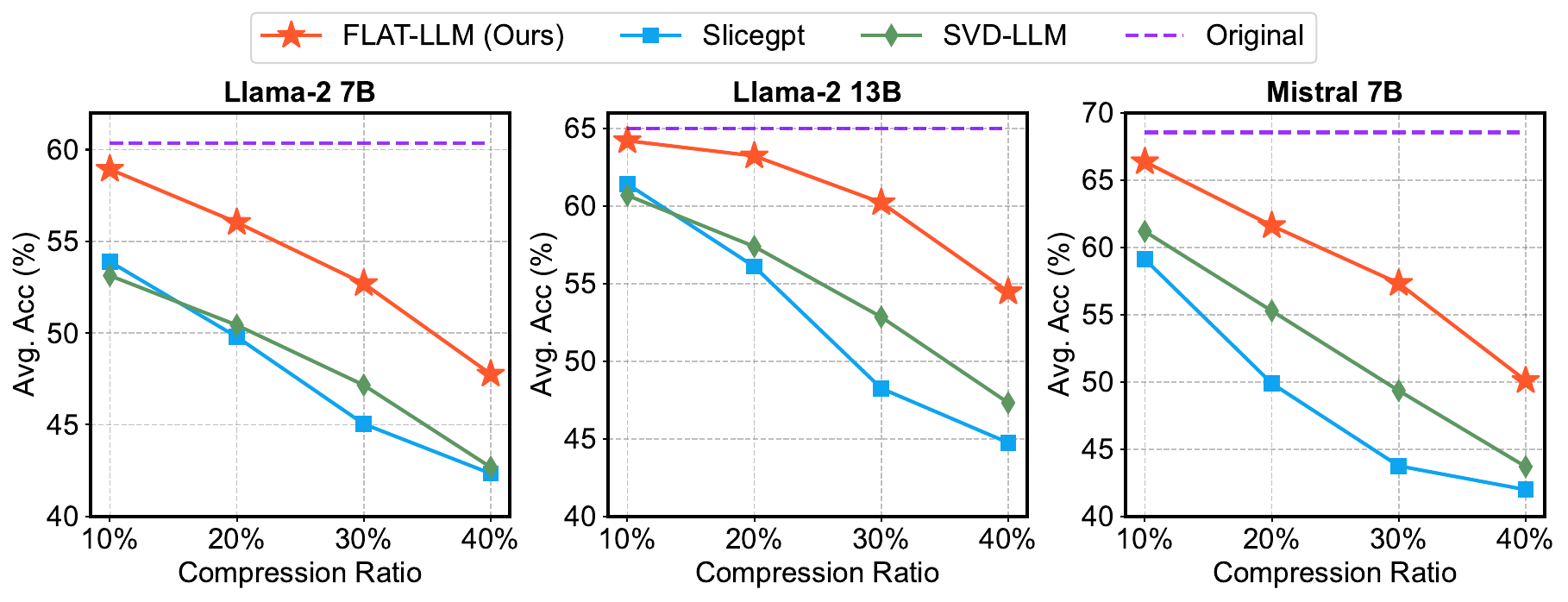}
    \caption{Comparison of zero-shot average accuracy versus compression ratio on various models.}
    \label{fig:acc_model}
\end{figure*}

\section{Additional Experimental Results}
\label{appendix:addition}

\subsection{Calibration Efficiency}
\label{appendix:cali_effi}
\begin{table}[htbp]
\centering
\caption{Comparison of calibration time and performance with other methods on LLAMA-2 13B at 20\% compression ratio.} 
\resizebox{.5\textwidth}{!}{
\begin{tabular}{c|c|c|c}
\hline
\textbf{Method}   & \textbf{Time} $\downarrow$ & \textbf{PPL} $\downarrow$  &\textbf{Avg. ACC} $\uparrow$            \\ \hline
SliceGPT \cite{ashkboosslicegpt}   & 0h35m     & 7.10      & 0.51 \\
SVD-LLM \cite{wang2024svd}    & 0h27m     & 7.69      & 0.56 \\
FLAP \cite{an2024fluctuation}       & 0h10m     & 5.90      & 0.57 \\

\rowcolor[HTML]{DAE8FC} 
\textbf{FLAT-LLM (ours)}    & 0h15m     & \textbf{5.55}       & \textbf{0.63} \\ \hline
\end{tabular}
}
\label{table:cali}
\end{table}
Table \ref{table:cali} compares the calibration time, perplexity (PPL), and average zero-shot accuracy of various compression methods at a 20\% compression ratio on Llama-2 13B. The results are collect from a single A100 GPU.
Among the evaluated methods, FLAT-LLM achieves the best overall performance, attaining the lowest perplexity and the highest average accuracy, while maintaining a moderate calibration time of 15 minutes. 
In contrast, SliceGPT and SVD-LLM exhibit $1.8-2.3\times$ longer calibration times and significantly $7-12\%$ accuracy drop. 
Although FLAP achieves the shortest calibration time, it suffers from a $6\%$ accuracy gap compared to our FLAT-LLM.
These results highlight that FLAT-LLM offers the best trade-off between calibration efficiency and compression performance, demonstrating high practical deployability.

\begin{figure*}[htbp]
    \centering

    \subfigure[Llama-3 8B]{%
        \includegraphics[width=0.45\linewidth]{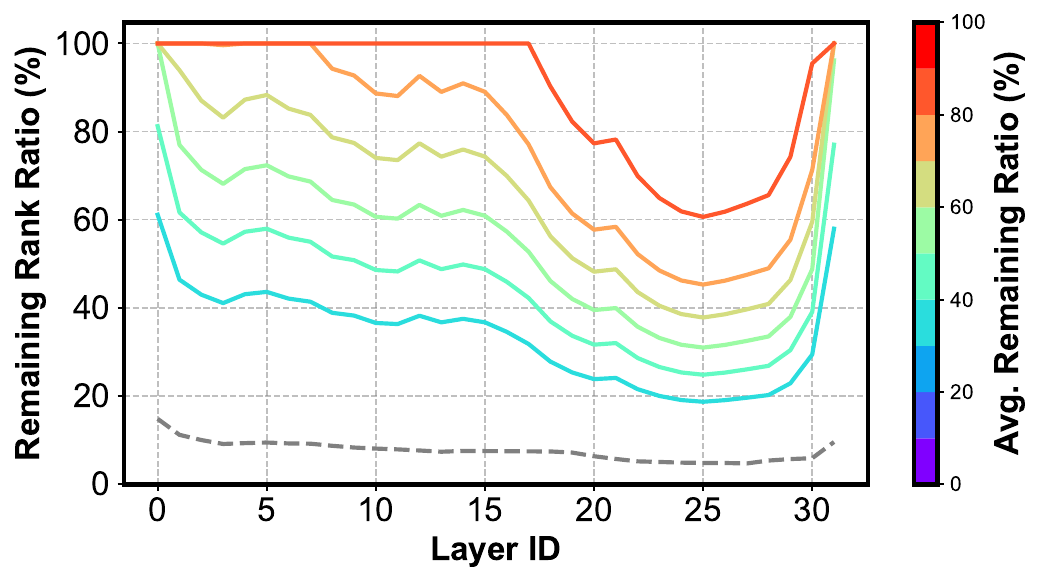}
    }
    \subfigure[Mistral-7B]{%
        \includegraphics[width=0.45\linewidth]{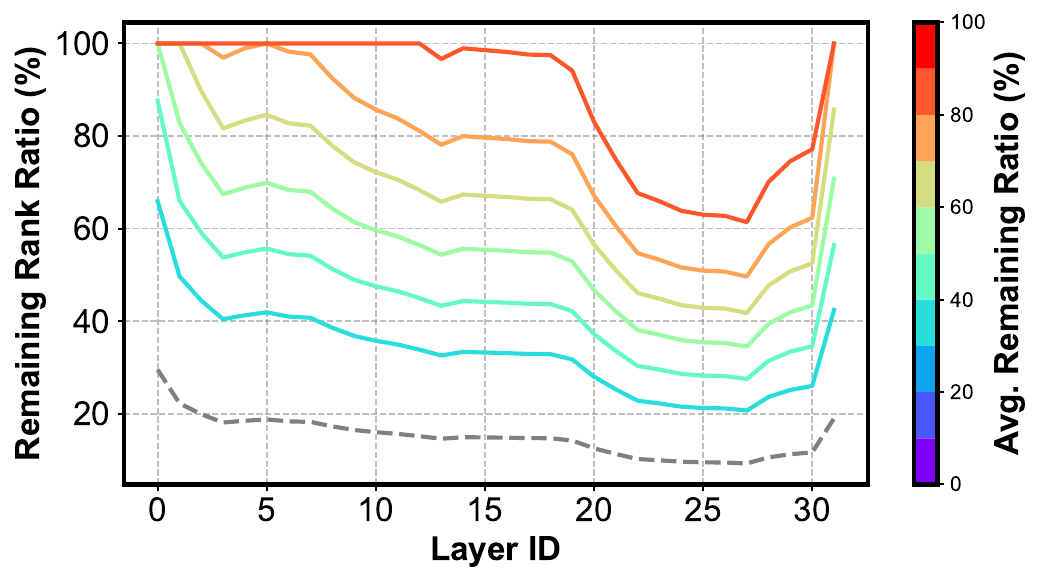}
    }
    
        \subfigure[Llama-2 13B]{%
        \includegraphics[width=0.45\linewidth]{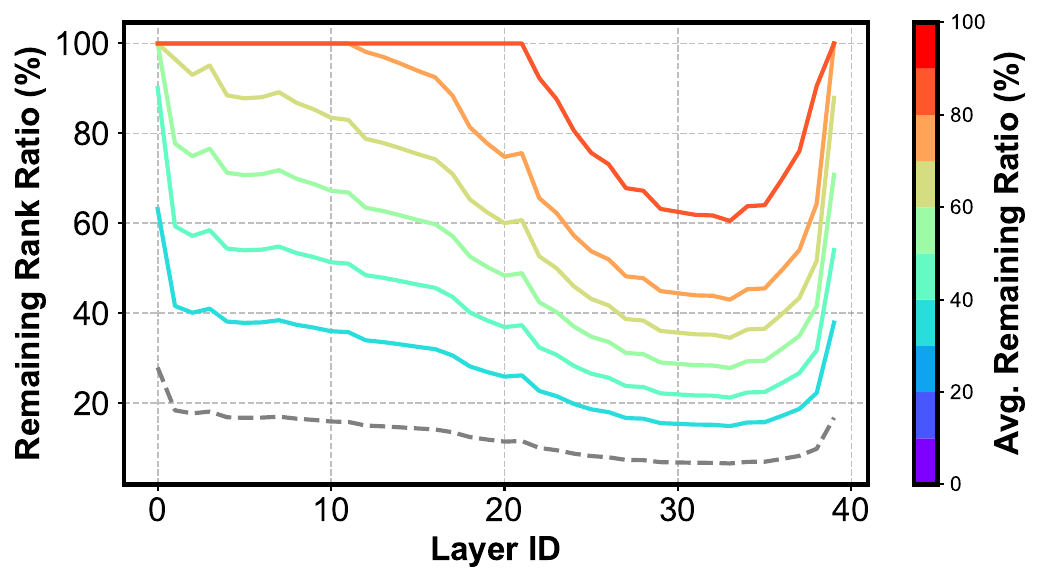}
    }
    \subfigure[Llama-2 70B]{%
        \includegraphics[width=0.45\linewidth]{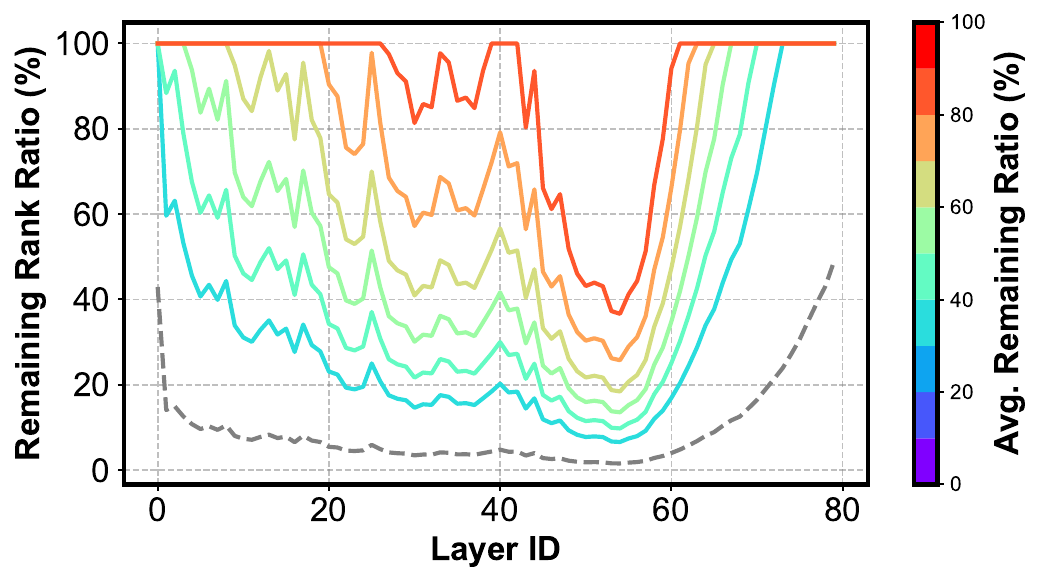}
    }

    \caption{Layer-wise sparsity score visualization across different LLM models using our IPRS algorithm.}
    \label{fig:app-IPRS}
\end{figure*}

\subsection{Additional Evaluation on Language Modeling}
\begin{table}[htbp]
\centering
\caption{Performance on LLaMA-3 8B evaluated on WikiText-2.}
\resizebox{.5\textwidth}{!}{
\begin{tabular}{c|c|c|c}
\hline
\textbf{Method} & \textbf{20\%} & \textbf{30\%} & \textbf{40\%} \\ \hline
Pivoting Factorization \cite{zhaopivoting} & 8.31 & 10.83 & 16.41 \\
LLM Surgeon \cite{vanllm}                  & 11.27 & 18.80 & 52.97 \\
\rowcolor[HTML]{DAE8FC} 
\textbf{FLAT-LLM (ours)}                   & \textbf{8.15} & \textbf{9.52} & \textbf{11.67} \\ \hline
\end{tabular}
}
\label{tab:other_baselines}
\end{table}
\label{appendix:ppl}
\textcolor{black}{We further compare our FLAT-LLM against recent structural pruning methods, including Pivoting Factorization \cite{zhaopivoting} and LLM Surgeon \cite{vanllm}. 
All methods are calibrated using 128 samples from WikiText-2, and perplexity is evaluated on the WikiText-2 test split. As shown in Table~\ref{tab:other_baselines}, FLAT-LLM consistently outperforms both Pivoting Factorization and LLM Surgeon on LLaMA-3 8B across various sparsity levels. 
In addition to strong performance, FLAT-LLM is significantly more efficient: it requires only 8 minutes of calibration on a single A100 40GB GPU, whereas LLM Surgeon takes approximately 4 hours using 8×A100 40GB GPUs when \( n_{\text{shots}} = 4 \).}

\subsection{Additional Evaluation on Zero-shot Downstream tasks}
\label{appendix:acc}
Figure~\ref{fig:acc_model} presents a comprehensive comparison of average zero-shot accuracy across compression ratios ranging from 10\% to 40\%, evaluated on three LLM models: LLaMA-2 7B, LLaMA-2 13B, and Mistral-7B. 
Calibration is performed using 128 samples with a sequence length of 4096 from the Alpaca dataset. The performance of the uncompressed dense model is shown as the upper bound (purple dashed line).
Our proposed method, FLAT-LLM, consistently outperforms prior low-rank decomposition approaches, including SlicedGPT and SVD-LLM, across all models and compression levels. 
Notably, FLAT-LLM maintains high performance with less than a $2\%$ accuracy drop at 10\% compression across all models. 
These results underscore the effectiveness and scalability of FLAT-LLM across varying models and compression regimes.

We further evaluate FLAT-LLM against additional importance-based baselines on five reasoning tasks, as shown in Table~\ref{table:acc_alpaca}. 
Following the setting of BlockPurner, we use 256 samples on Alpaca datasets for calibration. 
On average, FLAT-LLM incurs less than a 4\% accuracy drop when pruning Llama-2 7B at a 20\% compression ratio, and approximately a 6\% drop on Llama-2 13B at a 30\% compression ratio—significantly outperforming all baselines. Moreover, FLAT-LLM consistently achieves the highest accuracy across all benchmarks, demonstrating strong generalization across tasks.

\subsection{Illustration for IPRS Rank Distribution}
\label{appendix:IPRS}
Figure~\ref{fig:app-IPRS} presents the layer-wise remaining rank ratios produced by the IPRS algorithm across various LLM models, including Llama-3 8B, Llama-2 13B, 70B, and Mistral-7B. For each model, curves are shown under different target average remaining ratios, ranging from 30\% to 90\%. 
To compute the importance of each model, we use 128 samples with 4096 sequence length randomly selected from WikiText-2 dataset.
The gray dashed line represents the normalized importance scores $\mathbf{t}$ used to guide the adaptive rank allocation. 
Across all models, IPRS consistently preserves more capacity in layers deemed more important, while aggressively pruning less critical layers. 
Notably, the resulting rank distributions exhibit a U-shaped or skewed U-shaped pattern, allocating higher ranks to early, middle, or final layers depending on model-specific importance trends. 
Despite variations in architecture and depth, all models share a common pattern in which the input-adjacent and output-adjacent layers are more heavily preserved, reflecting their broader importance for information transformation and representation. 
This consistent behavior across diverse models highlights the robustness and generalization ability of the IPRS algorithm in learning effective, non-uniform rank allocations tailored to model-specific importance profiles.

\begin{figure}
    \centering
    \includegraphics[width=.9\linewidth]{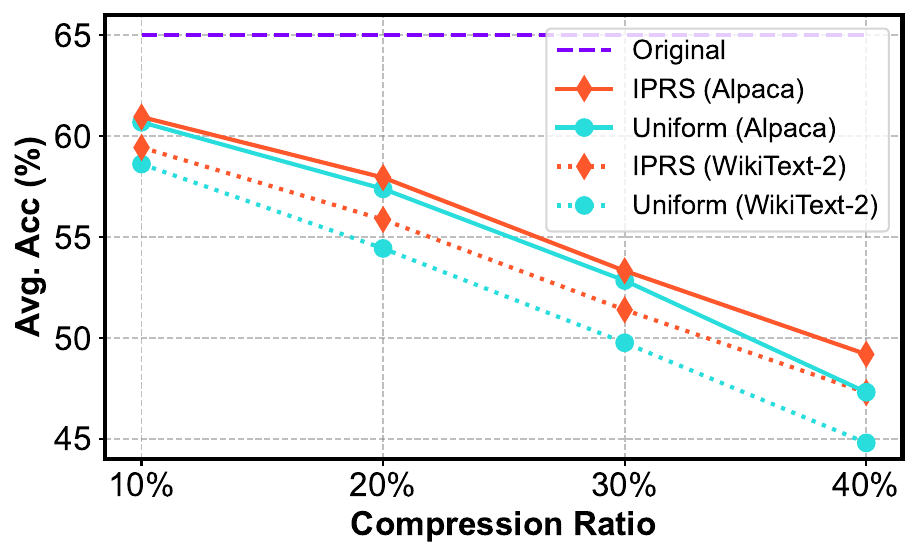}
    \caption{Comparison of average zero-shot accuracy across eight downstream datasets versus compression ratio using uniform rank and our IPRS algorithm on SVD-LLM.}
    \label{fig:iprs_svdllm}
\end{figure}

\subsection{IPRS on Other Decomposition Methods}
\label{appendix:IPRS2}
To evaluate the generality of our rank selection method, we also apply IPRS to SVD-LLM, as shown in Figure \ref{fig:iprs_svdllm}. While it leads to improvements in this setting, the average accuracy gain is modest, typically up to $2.5\%$. This is due to the distinct truncation loss patterns across layers: SVD-LLM exhibits relatively uniform truncation loss, whereas FLAT-LLM displays highly variable loss, amplifying the benefit of adaptive rank allocation. Overall, combining IPRS with our head-wise PCA-based compression in FLAT-LLM yields consistently superior performance, underscoring the complementary strengths of fine-grained PCA and importance-aware rank selection.

\subsection{Comparison with Other Rank-Adaptive Methods}
\label{appendix:modegpt}
\begin{table}[htbp]
\centering
\caption{Comparison of FLAT-LLM with IPRS and ModeGPT rank allocation strategies on LLaMA-3 8B evaluated on WikiText-2.}
\resizebox{.5\textwidth}{!}{
\begin{tabular}{c|c|c|c} \hline
\textbf{Method} & \textbf{30\%} & \textbf{40\%} & \textbf{50\%} \\ \hline
FLAT-LLM (w/ ModeGPT’s rank allocation, $\epsilon=0.2$) & 9.66 & 15.51 & 81.86 \\
FLAT-LLM (w/ ModeGPT’s rank allocation, $\epsilon=1$)   & 62.16 & 115.86 & 24.19 \\
\rowcolor[HTML]{DAE8FC} 
\textbf{FLAT-LLM (w/ our IPRS)}                         & \textbf{9.52} & \textbf{11.67} & \textbf{16.58} \\ \hline
\end{tabular}}
\label{tab:modegpt}
\end{table}

\textcolor{black}{We claim that our IPRS algorithm is entirely hyperparameter-free and achieves higher accuracy than MoDeGPT’s rank allocation method. In Table~\ref{tab:modegpt}, we apply both IPRS and MoDeGPT’s allocation strategies to FLAT-LLM and evaluate the resulting perplexity on the WikiText-2 test split. The results show that IPRS consistently outperforms MoDeGPT’s allocation method across all compression ratios and hyperparameter settings of~$\epsilon$. Furthermore, unlike MoDeGPT, our approach avoids the instability caused by the sensitivity of its hyperparameter~$\epsilon$.}

\end{document}